\documentclass[11pt,letterpaper]{article}
\usepackage[margin=1in]{geometry}
\usepackage[T1]{fontenc}
\usepackage[utf8]{inputenc}
\usepackage{lmodern}
\usepackage{microtype}
\usepackage{amsmath,amssymb,amsthm}
\usepackage{graphicx}
\usepackage{booktabs,array}
\usepackage{caption}
\usepackage{placeins}
\usepackage{titlesec}
\usepackage[authoryear,round]{natbib}
\let\cite\citet

\usepackage{xurl}
\usepackage[hidelinks]{hyperref}
\DeclareUnicodeCharacter{2212}{\ensuremath{-}}
\providecommand{\tightlist}{\setlength{\itemsep}{0pt}\setlength{\parskip}{0pt}}
\titleformat{\section}{\large\bfseries}{\thesection}{0.7em}{}
\titleformat{\subsection}{\normalsize\bfseries}{\thesubsection}{0.7em}{}
\newcounter{statement}
\newcounter{appendixlemma}[section]
\renewcommand{\theappendixlemma}{\thesection.\arabic{appendixlemma}}
\newcommand{\statementheading}[3]{%
  \refstepcounter{statement}\label{#2}%
  \textbf{#1 \thestatement#3.}}
\newcommand{\lemmaheading}[2]{%
  \refstepcounter{appendixlemma}\label{#1}%
  \textbf{Lemma \theappendixlemma#2.}}
\allowdisplaybreaks[1]
\hypersetup{
  pdftitle={OSCAR: Order-aware Sensitivity Correction for AI Rankings},
  pdfauthor={You Liu, Yue Liu, Quanchao Lu, Ergan Shang, Nick Shipilov},
  pdfsubject={OSCAR preprint}
}
\title{\LARGE\bfseries OSCAR: Order-aware Sensitivity Correction for AI Rankings}
\author{\normalsize You Liu\quad Yue Liu\quad Quanchao Lu\quad Ergan Shang\quad Nick Shipilov\\[0.5em]
\small\textit{Authors are listed in alphabetical order by surname.}}
\date{}
\begin{document}
\maketitle
% Generated from ../paper.md by generate.py. Formatting only.

\begin{abstract}

LLM judges evaluate candidate language models through pairwise comparisons of their responses. Judge-aware Bradley--Terry models aggregate pairwise comparisons by jointly estimating candidate scores and judge-specific sensitivities (how strongly each judge responds to quality differences), often interpreted as reliability. We show that this reading fails when judges prefer a display position. Even under balanced display orders or swap-and-average, omitting position attenuates fitted judge sensitivity by approximately the factor \(1-D^2\) near equal quality, where \(D\) is the judge\textquotesingle s net first-slot preference between equally good responses. We propose OSCAR, which jointly estimates candidate scores, judge sensitivities and judge-specific position, length and family effects, and provides confidence intervals for pairwise win probabilities that account for dependence within prompts. On held-out data, observed attenuation closely matches the prediction \(1-D^2\) (correlation of log ratios 0.99). Across four datasets, position adjustment lowers held-out log loss by 14--23\% and changes which judges would be dropped for low sensitivity on three datasets. On MT-Bench, it ranks GPT-4 first instead of claude-v1, matching human votes. Simulations with known parameters show improved recovery of judge sensitivities and candidate rankings, with 94.4--95.2\% confidence interval coverage. Low estimated sensitivity alone is therefore not evidence of an unreliable judge.

\end{abstract}

\section{Introduction}
\label{sec:1}

Pairwise comparison by LLM judges has become a standard way to evaluate language models, which we call candidates \citep{zheng2023}. An LLM judge compares two responses to the same prompt and selects the better one. These judgments, often collected from several judges, are then aggregated into candidate scores and rankings with a Bradley--Terry model \citep{bradley1952,chiang2024}. Because judges differ in how reliably they separate strong responses from weak ones, recent judge-aware models give each judge its own sensitivity, a slope that scales how strongly a quality difference moves that judge\textquotesingle s preferences \citep{xu2026a,qian2026}. Fitted sensitivities determine judges\textquotesingle{} influence in ranking aggregation \citep{xu2026a}. More broadly, reliability estimates are used to select and weight judges \citep{li2025}, while groups of diverse judges can replace a single evaluator \citep{verga2024}.

\begin{figure}[!htbp]
\centering
\begin{minipage}[t]{0.65\linewidth}\vspace{0pt}\centering
{\small (a) Slot-A win rate minus slot-B win rate for In-House judges\par}
\smallskip
\includegraphics[width=\linewidth]{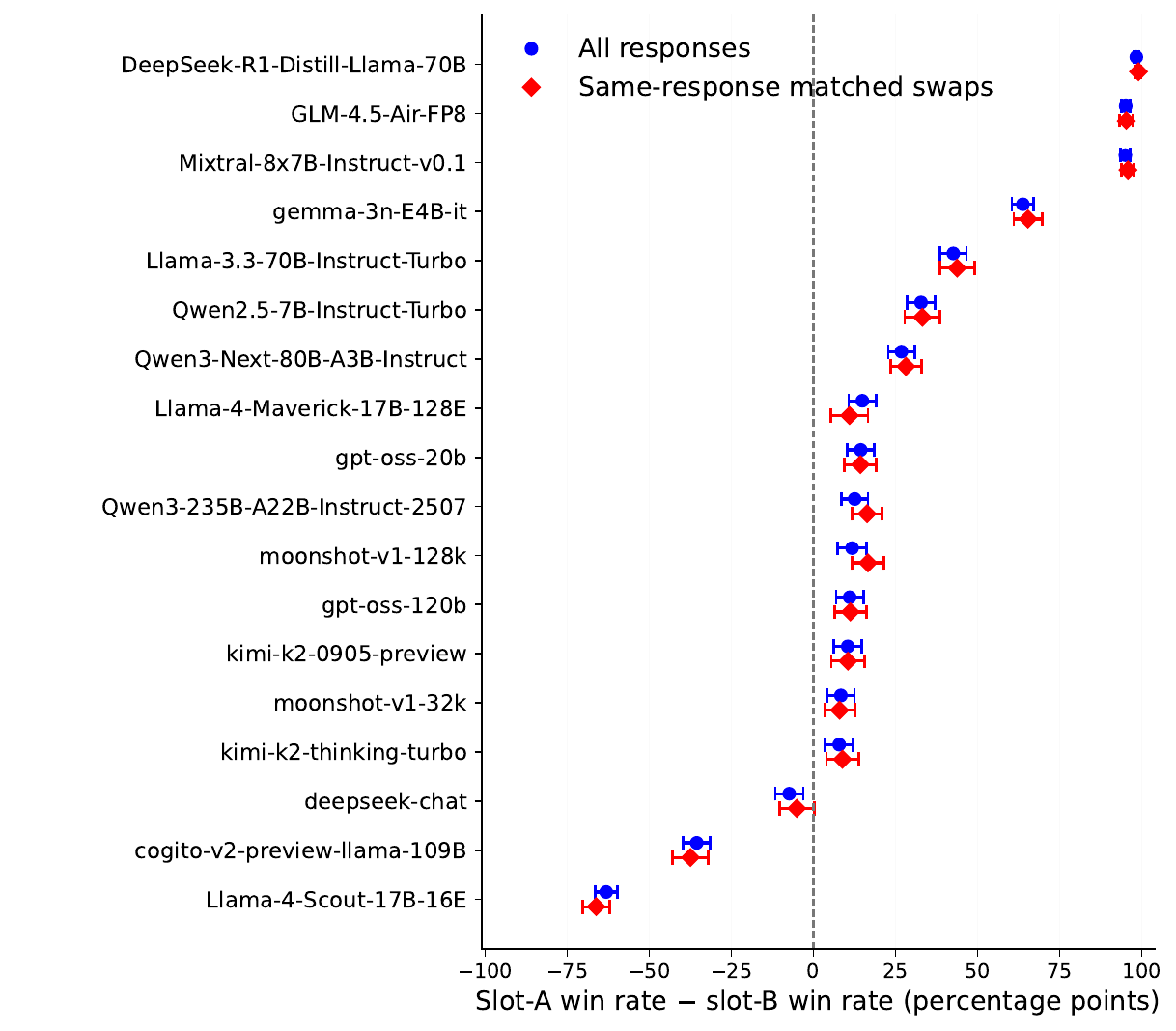}
\end{minipage}\hfill
\begin{minipage}[t]{0.33\linewidth}\vspace{0pt}\centering
{\small (b) Exact calculation with known true sensitivity \(\gamma^* = 1\)\par}
\smallskip
\includegraphics[width=\linewidth]{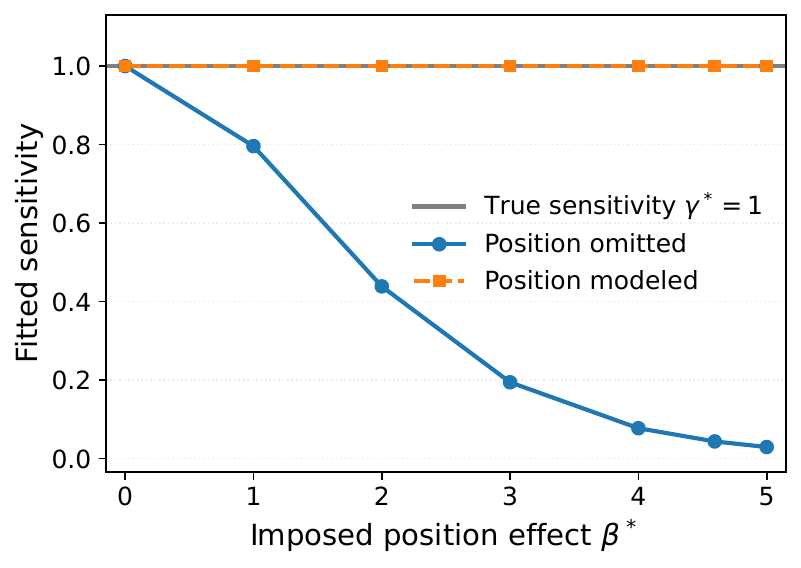}
\par\medskip
{\small (c) Held-out sensitivity attenuation with unknown true sensitivities\par}
\smallskip
\includegraphics[width=\linewidth]{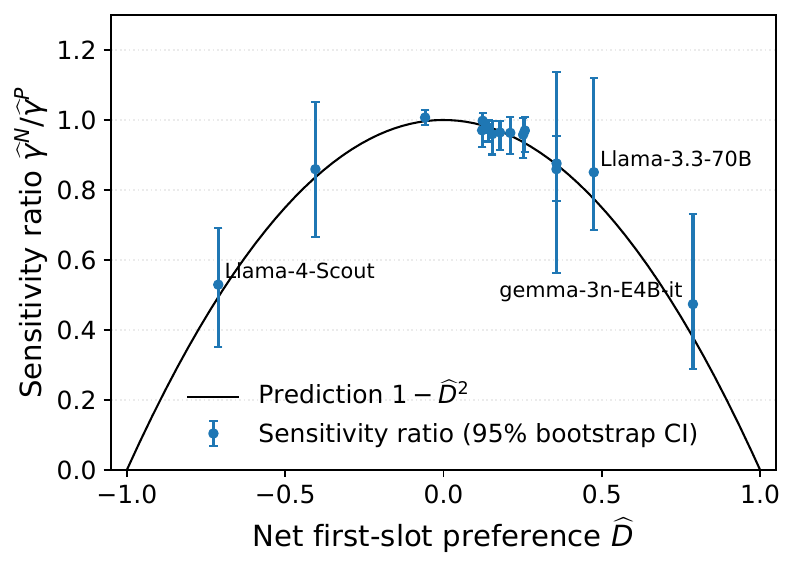}
\end{minipage}
\caption{Position bias masquerades as low judge sensitivity. \textbf{(a)} Slot-A win rate minus slot-B win rate for the 18 In-House judges, over all responses (blue) and over same-response matched swaps (red), with 95\% confidence intervals accounting for dependence within questions (Appendix~\ref{app:C-4}). Positive values mean a preference for the first slot. DeepSeek-R1-Distill-Llama-70B picks the first-shown response in more than 99\% of judgments on matched swaps. \textbf{(b)} Exact population calculation (Equation~\eqref{eq:6}) with known true sensitivity \(\gamma^*=1\). The fitted sensitivity falls from 1 to 0.08 as the position effect increases from \(\beta^*=0\) to \(\beta^*=4\) when position is omitted (blue), and remains at 1 when position is modeled (orange). \textbf{(c)} Held-out attenuation among 15 In-House judges with finite, positive sensitivity ratios. Position effects \(\hat\beta_k\) and candidate scores are estimated on 594 calibration prompts. On 198 separate prompts, sensitivities are refitted with these scores fixed, without a position term (\(\widehat\gamma^{\mathrm N}_k\)) and with the position effect \(\hat\beta_k\) fixed (\(\widehat\gamma^{\mathrm P}_k\), position-adjusted). Points show each judge\textquotesingle s fitted sensitivity ratio \(\widehat\gamma^{\mathrm N}_k/\widehat\gamma^{\mathrm P}_k\) matches the attenuation prediction \(1-\widehat D_k^2\). Bars are 95\% bootstrap intervals. GLM-4.5-Air has a zero sensitivity ratio, while DeepSeek-R1-Distill-Llama-70B and Mixtral-8x7B have zero adjusted sensitivities and therefore undefined ratios, so these three judges are excluded.}
\label{fig:1}
\end{figure}

However, consider DeepSeek-R1-Distill-Llama-70B on the In-House dataset released by \cite{xu2026a}. Shown the same two responses in both orders, it chooses the first response in more than 99\% of judgments on matched swaps (Figure~\ref{fig:1}(a)), and ranks 17th of 18 judges by fitted sensitivity in a fit that omits position. More broadly, among the eight In-House judges in \cite[Table 2]{xu2026a}, published sensitivity falls almost monotonically as the all-response absolute A/B imbalance grows (Spearman correlation −0.98, Table~\ref{tab:C2}). These observations raise a question about how estimated sensitivity should be interpreted: is such a judge unreliable, or is the model mistaking position bias for low sensitivity?

Omitting position can attenuate fitted judge sensitivity. In the calculation in Appendix~\ref{app:A-3}, the judge has true sensitivity 1 and position effect \(\beta^*=1\). The higher-quality response is displayed first in 40\% of comparisons and second in 60\%. Fitting the exact preference probabilities gives sensitivity 0.528 without a position term and recovers 1 when position is included.

This attenuation can persist even under balanced presentation or swap-and-average. Balanced presentation assigns each response pair to its two display orders with equal probability. In a swap-and-average protocol \citep{wang2024,qian2026}, the same judge evaluates the same two responses in both orders, and the resulting preference scores are averaged after aligning them to the same response. Under either protocol, the expected outcome at quality gap \(x\) is the average of the judge\textquotesingle s response curves for the two slots, \(m(x)=\{\sigma(\gamma^* x+\beta^*)+\sigma(\gamma^* x-\beta^*)\}/2\). Here, \(\sigma\) is the logistic function, \(\gamma^*\) is the true sensitivity and \(\beta^*\) is the true position effect. Fitting \(\sigma(\gamma x)\) to this mean by minimizing expected log loss gives the population fit \(\gamma^\dagger\). With quality scores held fixed and \(\beta^*\ne0\), the population fit has \(0<\gamma^\dagger<\gamma^*\), so omitting position attenuates fitted judge sensitivity under either protocol. When comparisons are concentrated near equal quality, we show \(\gamma^\dagger\approx\gamma^*(1-(D^*)^2)\), where \(D^*=\tanh(\beta^*/2)\) is the judge\textquotesingle s net first-slot preference between equally good responses (Section~\ref{sec:4-1}). For example, a judge that selects the first of two equally good responses 90\% of the time (\(D^*=0.8\)) has its fitted sensitivity reduced to about 36\% of the true value. In real data, swap-and-average leaves the scale-invariant sensitivity at 54\% of the position-adjusted value for Llama-4-Scout and 71\% for gemma-3n-E4B on In-House (Section~\ref{sec:6-2}).

Figure~\ref{fig:1}(b) illustrates the attenuation effect in an exact population calculation with true sensitivity fixed at 1: the fitted sensitivity falls from 1 to 0.08 as the position effect grows to \(\beta^* = 4\). Figure~\ref{fig:1}(c) shows that the same attenuation occurs in real judges, where the true sensitivities are unknown. We test the prediction on held-out In-House data: position effects and candidate scores are estimated on one set of prompts, and each judge\textquotesingle s sensitivity is refitted on disjoint prompts with and without the estimated position effect (Section~\ref{sec:6-2}), giving a naive sensitivity \(\widehat\gamma^{\mathrm N}_k\) and a position-adjusted sensitivity \(\widehat\gamma^{\mathrm P}_k\). Among the 15 judges with finite, positive sensitivity ratios, 14 have higher fitted sensitivity once the position effect is included (Llama-4-Scout: 0.36 to 0.67), and the ratio \(\widehat\gamma^{\mathrm N}_k/\widehat\gamma^{\mathrm P}_k\) closely follows the prediction \(1-\widehat D_k^2\) (the correlation between log observed and log predicted ratios is 0.99, and the geometric-mean observed-to-predicted ratio is 1.03, 95\% CI 0.98--1.08).

We propose OSCAR (Order-aware Sensitivity Correction for AI Rankings), which combines the judge-aware Bradley--Terry structure with judge-specific effects for display position, response length and whether the candidate comes from the judge\textquotesingle s own model family. We also provide cluster-robust confidence intervals for pairwise win probabilities that account for dependence among judgments of the same prompt. Across four datasets, position adjustment improves held-out prediction. It also changes the bottom-three judge set on three of the four datasets and which model tops a leaderboard. For example, on UltraFeedback, Mixtral-8x7B is the second least sensitive of 20 judges without a position term and ranks 15th with one (Figure~\ref{fig:2}). On MT-Bench, the fit without position ranks claude-v1 first, whereas the position-adjusted fit and OSCAR rank GPT-4 first and match the ranking from human votes (Section~\ref{sec:6-4}). Simulations with known parameters evaluate judge sensitivity recovery, candidate ranking recovery and confidence interval coverage. Our contributions are:

\begin{enumerate}
\tightlist
\item
  \textbf{Mechanism.} We show that omitting position changes the interpretation of estimated judge sensitivity, with attenuation by a factor of approximately \(1-D^2\) (when quality gaps are concentrated near zero) that persists even under balanced presentation and swap-and-average. Low fitted sensitivity alone therefore does not imply a weak ability to distinguish response quality.
\item
  \textbf{Model.} We introduce OSCAR and jointly estimate candidate scores, judge sensitivities and judge-specific position, length and family effects using fractional-response quasi-likelihood, with ties scored as one half. We state local identification conditions and show when changes in candidate scores can be exactly offset by changes in family effects.
\item
  \textbf{Inference.} We construct cluster-robust confidence intervals for pairwise win probabilities that account for dependence among judgments of the same prompt. Under the conditions of Theorem~\ref{thm:cluster}, the probability that an interval contains the true win probability approaches its stated confidence level as the number of independent prompts grows.
\item
  \textbf{Evidence.} A held-out test in real data is consistent with the predicted attenuation \(1-D^2\) across judges. Across four datasets, position adjustment improves held-out prediction and, on three of them, changes which judges a sensitivity-based screen would drop. On MT-Bench, position adjustment changes the top-ranked candidate from claude-v1 to GPT-4, and the adjusted ranking matches human votes. In simulations, OSCAR improves sensitivity and ranking recovery, and its 95\% confidence intervals achieve 94.4--95.2\% coverage.
\end{enumerate}

\section{Related Work}
\label{sec:2}

\textbf{Judge-aware aggregation.} Bradley--Terry models aggregate pairwise judgments into candidate scores \citep{bradley1952,chiang2024}. Crowd-BT introduces annotator-specific reliability \citep{chen2013}. \cite{bugakova2019} show that Crowd-BT can treat biased workers as random responders and introduce factorBT, a mixture of quality-based and feature-based preferences. OSCAR models quality sensitivity and bias effects within a single logistic predictor and quantifies how omitting position changes the fitted quality slope. Recent work on LLM evaluation combines judgments from multiple judges \citep{verga2024}, learns judge-specific sensitivities \citep{xu2026a,qian2026}, separates sensitivity from residual judge disagreement \citep{yu2026}, and develops more general probabilistic comparisons and uncertainty estimates \citep{fathullah2025}. \cite{qian2026} average the probabilities of choosing the same candidate across both display orders before fitting their Bradley--Terry models. We show why these averaged preferences can still yield an attenuated sensitivity estimate (Section~\ref{sec:4-1}).

\textbf{Bias adjustment and identification.} Position, verbosity and self-preference effects are documented across LLM judges \citep{zheng2023,panickssery2024,shi2025}. Existing adjustments include evaluating both orders \citep{wang2024} and controlling for length and style covariates \citep{dubois2024,li2024}. Order effects and ties also have established extensions in classical paired-comparison models \citep{davidson1970,davidson1977,turner2012}. Identifying candidate quality separately from bias requires comparisons that distinguish their effects. \cite{xu2026b} show that this separation can fail when content covariates are confounded with latent quality. We give a local identification condition for OSCAR and show a two-family setting in which changes in candidate scores can be exactly offset by changes in family effects, leaving all predicted preferences unchanged (Section~\ref{sec:4-3}).

\textbf{Statistical foundations.} In logistic models, averaging over one predictor can change the fitted coefficient of another, even when the two are independent. This is known as logistic non-collapsibility \citep{gail1984,zeger1988,neuhaus1993}. We derive the exact factor \(1-D^2\) for the log-odds slope at equal quality (Section~\ref{sec:4-1}) and a first-order approximation to how omitted bias terms change jointly estimated candidate scores and judge sensitivities (Theorem~\ref{thm:omitted-shifts}). Estimation uses fractional-response quasi-likelihood, which applies the same log-loss objective to binary outcomes and ties scored as one half \citep{papke1996}. Following \cite{white1982}, we describe fits that omit relevant effects by the parameter values that minimize expected loss. Prompt-cluster inference accounts for dependence among judgments of the same prompt \citep{cameron2015,miller2024}, under the conditions of Theorem~\ref{thm:cluster}.

\section{Judge-Sensitive Ranking with Order Effects}
\label{sec:3}

\subsection{Observation}
\label{sec:3-1}

We observe \(N\) pairwise judgments of \(M\) candidates by \(K\) LLM judges, grouped into \(G\) prompt clusters. Each cluster contains the judgments associated with one prompt. For \(t=1,\ldots,N\), record \(t\) contains its cluster \(g(t)\), the candidates \(a_t\) and \(b_t\) actually shown in slots A and B, the judge \(k_t\), a signed length contrast \(L_t\) and a family contrast \(F_t\). We encode the outcome (A, B or a tie) as the slot-A score

\begin{equation}
\label{eq:1}
Y_t = \mathbf{1}\{A\} + \tfrac{1}{2}\,\mathbf{1}\{\text{tie}\}
\end{equation}

Its conditional mean, \(p_t=P(A)+\tfrac12P(\text{tie})\), is the tie-adjusted slot-A win probability used throughout the paper. Records keep the displayed order. Relabeling the preferred response as A would erase the slot information that the position coefficient measures (Section~\ref{sec:4}).

\subsection{Model}
\label{sec:3-2}

Let \(s_i\) be the score of candidate \(i\) and \(\gamma_k > 0\) the sensitivity of judge \(k\). Our model OSCAR is

\begin{equation}
\label{eq:2}
p_t = \sigma(\eta_t), \qquad \eta_t = \gamma_{k_t}(s_{a_t} - s_{b_t}) + \beta_{\mathrm{pos},k_t} + \beta_{\mathrm{len},k_t}L_t + \beta_{\mathrm{fam},k_t}F_t
\end{equation}

where \(\sigma(x)=(1+e^{-x})^{-1}\).

A positive \(\beta_{\mathrm{pos},k}\) favors slot A. For a comparison judged by \(k\) with \(s_{a_t}=s_{b_t}\), \(L_t=0\) and \(F_t=0\), Equation~\eqref{eq:2} gives \(p_t = \sigma(\beta_{\mathrm{pos},k})\), so judge \(k\)\textquotesingle s net first-slot preference is

\[
D_k = P(A) - P(B) = 2p_t - 1 = 2\sigma(\beta_{\mathrm{pos},k}) - 1 = \sigma(\beta_{\mathrm{pos},k}) - \sigma(-\beta_{\mathrm{pos},k}) = \tanh(\beta_{\mathrm{pos},k}/2),
\]

which ranges from −1 (always choose slot B) to 1 (always choose slot A). Section~\ref{sec:4-1} shows that, under balanced display orders and near equal quality, omitting position reduces fitted sensitivity by approximately the factor \(1-D_k^2\).

The length contrast is

\[
L_t = \log\{(\ell_{A,t}+1)/(\ell_{B,t}+1)\}/s_L,
\]

where \(\ell_{A,t}\) and \(\ell_{B,t}\) are the lengths of the two responses in Unicode characters and \(s_L\) is the dataset-wide empirical standard deviation of this log ratio, computed with divisor \(N\), so that \(L_t\) is measured in standard-deviation units. It is zero at equal length and changes sign under a swap.

The family contrast is

\[
F_t = \mathbf{1}\{\mathrm{fam}(a_t) = \mathrm{fam}(k_t)\} - \mathbf{1}\{\mathrm{fam}(b_t) = \mathrm{fam}(k_t)\}
\]

where \(\mathrm{fam}(\cdot)\) maps a candidate or judge to its model family (Appendix~\ref{app:B-2}). The contrast marks whether a response comes from the judge\textquotesingle s model family. It equals 1 when only the response in slot A comes from the judge\textquotesingle s own model family, −1 when only the response in slot B does, and 0 when both or neither do.

\textbf{Remark.} We name the nested versions by their terms. J keeps only \(\gamma_k(s_a - s_b)\), as in judge-aware Bradley--Terry models \citep{xu2026a}. J+P, J+L and J+F add the position, length or family term, respectively, and OSCAR includes all three.

\subsection{Pairwise win probability}
\label{sec:3-3}

The parameters are not identifiable on their own: adding a constant to every score leaves every prediction unchanged, and so does multiplying all scores by \(c>0\) while dividing all sensitivities by \(c\). We remove this freedom by imposing two constraints: \(\sum_i s_i = 0\) and \(\gamma_{\mathrm{ref}} = 1\) for a reference judge. Individual scores and sensitivities depend on the location and scale conventions. We therefore define a comparison target that is invariant under these transformations:

\begin{equation}
\label{eq:3}
\bar\gamma = \frac{1}{K}\sum_{k=1}^{K}\gamma_k, \qquad h_{ij} = \bar\gamma(s_i - s_j), \qquad q_{ij} = \sigma(h_{ij})
\end{equation}

where \(\bar\gamma\) is the average sensitivity, so that \(h_{ij}\) is the log-odds with which a judge of average sensitivity prefers \(i\) to \(j\). For candidate comparisons, we report \(q_{ij}\), the tie-adjusted win probability for a judge with average sensitivity \(\bar\gamma\), evaluated with the position, length and family contributions set to zero. This provides a common reference for comparing candidates. The target \(q_{ij}\) is the quantity whose confidence intervals we report (Section~\ref{sec:5}) and whose error we measure in simulation (Section~\ref{sec:7}). For held-out prediction in Section~\ref{sec:6-3}, we retain the bias terms and use Equation~\eqref{eq:2}.

\subsection{Estimation}
\label{sec:3-4}

We estimate the model by minimizing log loss with a ridge penalty of strength \(\lambda\):

\begin{equation}
\label{eq:4}
\mathcal L_N(\theta) = \sum_{t=1}^{N}\left\{\log\left(1 + e^{\eta_t}\right) - Y_t\eta_t\right\} + \frac{\lambda}{2}\lVert\theta\rVert^2
\end{equation}

over the free coordinates

\[
\theta = (u, v, b), \qquad s = Cu, \qquad \gamma_k = \exp(v_k)\ (k \neq \mathrm{ref}), \qquad \gamma_{\mathrm{ref}} = 1.
\]

The matrix \(C\in\mathbb R^{M\times(M-1)}\) has orthogonal unit-length columns whose entries sum to zero, with \(C^\top C=I\) and \(C^\top\mathbf 1=0\). Its construction is given in Appendix~\ref{app:B-1}. Write \(C_i\) for row \(i\) of \(C\), and let \(b\) collect the retained bias coefficients. Thus \(\theta\) collects the free score coordinates, non-reference log-sensitivities and retained bias coefficients. The parametrization enforces \(\sum_i s_i=0\), \(\gamma_{\mathrm{ref}}=1\) and positive sensitivities. For binary outcomes, the unpenalized part of Equation~\eqref{eq:4} is the Bernoulli negative log-likelihood. With tie-half outcomes, it is a fractional quasi-likelihood \citep{papke1996}.

For \(\lambda>0\), the ridge guarantees that a finite minimizer exists. When different parameter values give identical predictions, however, the penalty rather than the data selects among them, so the ridge does not resolve non-identification. After removing uninformative covariate columns, we check local identification through the full column rank of the predictor Jacobian \(R\), whose \(t\)th row is \((\nabla_\theta\eta_t)^\top\). Appendix~\ref{app:B} gives the parametrization, column-selection rules, optimization settings and numerical checks.

\textbf{Remark.} Throughout the paper, we use a superscript \(*\) for the true parameter, a dagger \(\dagger\) for a population minimizer, which minimizes expected loss, and a hat \(\widehat{}\) for a sample estimate. Under misspecification, the population minimizer is the pseudo-true parameter of the fitted model.

\section{How Omitted Order Distorts Sensitivity}
\label{sec:4}

This section derives how omitting position biases fitted sensitivities. Section~\ref{sec:4-1} obtains the attenuation for one judge with the scores fixed, Section~\ref{sec:4-2} allows scores and sensitivities to be estimated jointly, and Section~\ref{sec:4-3} states what the data must contain for the added terms to be identified.

\subsection{Sensitivity attenuation under balanced display orders}
\label{sec:4-1}

Consider one judge with true sensitivity \(\gamma^*>0\) and position intercept \(\beta^*\). Fix a response whose signed quality difference from the other response is \(x\). Set the length and family contrasts to zero. When this response is shown first or second, its expected tie-adjusted preference score is \(\sigma(\gamma^* x+\beta^*)\) or \(\sigma(\gamma^* x-\beta^*)\), respectively, where \(\beta^*=\beta_{\mathrm{pos}}^*\). If the two orders are equally likely at each quality gap, the order-averaged mean is

\begin{equation}
\label{eq:5}
\begin{aligned}
m(x)&=\tfrac{1}{2}\left\{\sigma(\gamma^* x + \beta^*) + \sigma(\gamma^* x - \beta^*)\right\},\\
\left.\frac{d\,\mathrm{logit}\,m(x)}{dx}\right|_{x=0}
&= \gamma^*\,\mathrm{sech}^2(\beta^*/2) = \gamma^*\,(1 - (D^*)^2)
\end{aligned}
\end{equation}

Here, \(\mathrm{logit}(p)=\log\{p/(1-p)\}\), and \(D^* = \tanh(\beta^*/2)\) is the judge\textquotesingle s net first-slot preference (Section~\ref{sec:3-2}). Averaging over orders removes the directional position preference: when the two responses are equally good (\(x = 0\)), the averaged curve gives \(m(0) = \tfrac{1}{2}\{\sigma(\beta^*) + \sigma(-\beta^*)\} = 1/2\), the same as for a judge without position preference. More generally, \(m(-x) = 1 - m(x)\).

But averaging does not remove sensitivity attenuation. Sensitivity measures how strongly a judge\textquotesingle s preference responds to quality differences, on the log-odds scale. Without position preference, \(\mathrm{logit}\{\sigma(\gamma^* x)\}=\gamma^* x\), so each unit of quality advantage increases the log-odds of being preferred by \(\gamma^*\). For the averaged curve, \(\mathrm{logit}\{m(x)\}\) is no longer linear: its slope at \(x=0\) is \(\gamma^*(1-(D^*)^2)\). Thus, near equal quality, the averaged preferences respond less strongly to quality differences. Swap-and-average retains this attenuation because it fits the same order-averaged mean.

A separate result concerns fitting a single logistic slope across multiple quality gaps. With the gaps fixed and not all zero, and with \(\beta^*\neq0\), the omitted-position population fit \(\gamma^\dagger\) satisfies \(0<\gamma^\dagger<\gamma^*\) (Lemma~\ref{lem:A2}, Appendix~\ref{app:A-3}). When these gaps concentrate near zero, \(\gamma^\dagger\approx\gamma^*(1-(D^*)^2)\) (Lemma~\ref{lem:A3}).

\textbf{Remark.} Equation~\eqref{eq:5} describes the slope at zero quality gap \(x = 0\). To illustrate the fitted slope across a range of gaps, Figure~\ref{fig:1}(b) uses true sensitivity \(\gamma^*=1\) and 401 equally spaced, equally weighted values \(x_j\) on \([-1,1]\). With position effect \(\beta^*\) omitted, the population fit is

\begin{equation}
\label{eq:6}
\begin{aligned}
\gamma^{\mathrm N,\dagger}(\beta^*)
&= \arg\min_{\gamma\geq0}\ \sum_{j=1}^{401}\left\{\log\left(1 + e^{\gamma x_j}\right) - m(x_j)\, \gamma x_j\right\},\\
m(x_j)
&= \tfrac{1}{2}\left\{\sigma(x_j + \beta^*) + \sigma(x_j - \beta^*)\right\}.
\end{aligned}
\end{equation}

Including position recovers \(\gamma^{\mathrm P,\dagger}(\beta^*)=1\). Omitting it reduces \(\gamma^{\mathrm N,\dagger}(\beta^*)\) from 1 at \(\beta^*=0\) to approximately 0.077 at \(\beta^*=4\). Appendix~\ref{app:A-2} derives Equation~\eqref{eq:5}, and Appendix~\ref{app:A-3} gives the calculation behind Equation~\eqref{eq:6} and Figure~\ref{fig:1}(b).

\subsection{Joint estimation of scores and sensitivities}
\label{sec:4-2}

Equation~\eqref{eq:5} holds the candidate scores fixed. When scores and sensitivities are estimated jointly, an omitted term can move both. Let \(\xi=(u,v)\) collect the free score and log-sensitivity coordinates, and let \(\xi^*=(u^*,v^*)\) denote their true values. Write \(\eta^{\mathrm J}(\xi)=\gamma_k(s_a-s_b)\) for the predictor of J (Section~\ref{sec:3-2}), which has judge sensitivities but no bias terms. We study the family of true means

\[
\mu_\varepsilon=\sigma\{\eta^{\mathrm J}(\xi^*)+z^\top\varepsilon\},
\]

where \(z\) contains the omitted covariates and \(\varepsilon\) contains their coefficients. At \(\varepsilon=0\), these effects are absent. For example, if only position is omitted, \(z\) is the vector of judge indicators and \(\varepsilon\) contains the judges\textquotesingle{} position coefficients. Define the unpenalized population loss and its global pseudo-true minimizer by

\[
\mathcal L(\xi,\varepsilon)
=E\left[
\log\{1+e^{\eta^{\mathrm J}(\xi)}\}
-\mu_\varepsilon\eta^{\mathrm J}(\xi)
\right],
\qquad
\xi^\dagger(\varepsilon)=\arg\min_\xi\mathcal L(\xi,\varepsilon).
\]

Taking expectations over comparison records according to their population frequencies, define

\[
\begin{aligned}
w^* &= \sigma\{\eta^{\mathrm J}(\xi^*)\}\left[1 - \sigma\{\eta^{\mathrm J}(\xi^*)\}\right],
& r^* &= \nabla_\xi\eta^{\mathrm J}(\xi^*),\\
H^* &= E[w^* r^* (r^*)^\top],
& B^* &= E[w^* r^* z^\top]
\end{aligned}
\]

The following theorem shows how far \(\xi^\dagger(\varepsilon)\) moves from \(\xi^*\) when the omitted effects \(\varepsilon\) are small.

\statementheading{Theorem}{thm:omitted-shifts}{ (Parameter shifts from omitted covariates)} Consider the J model under the constraints \(\sum_i s_i=0\) and \(\gamma_{\mathrm{ref}}=1\) with fixed finite sets of candidates and judges and a fixed distribution of ordered combinations \((a,b,k)\) of first-slot candidate, second-slot candidate and judge. Let \(\xi^*\) be finite, and let the omitted-covariate vector \(z\) be finite almost surely. Suppose the population loss gradient \(\nabla_\xi \mathcal L\) is twice continuously differentiable near \((\xi^*,0)\), differentiation under the expectation is justified, and \(H^*\succ0\). For all sufficiently small \(\varepsilon\), the global minimizer \(\xi^\dagger(\varepsilon)\) exists, is unique and satisfies

\begin{equation}
\label{eq:7}
\xi^\dagger(\varepsilon) - \xi^* = (H^*)^{-1}B^*\,\varepsilon + O(\lVert\varepsilon\rVert^2)
\end{equation}

The matrix \(B^*\) determines which omitted effects contribute at first order.

\statementheading{Corollary}{cor:balanced}{ (First-order protection from balanced presentation)} Under the conditions of Theorem~\ref{thm:omitted-shifts}, suppose the design is invariant under \((a,b,L,F,k)\mapsto(b,a,-L,-F,k)\). Then the position columns of \(B^*\) vanish. When position is the only omitted effect, \(\xi^\dagger(\varepsilon)-\xi^*=O(\lVert\varepsilon\rVert^2)\).

Balanced presentation eliminates the first-order parameter shift caused by omitted position. Second-order shifts can remain. This is consistent with the expansion of Equation~\eqref{eq:5}, \(\mathrm{sech}^2(\beta^*/2) = 1 - (\beta^*)^2/4 + O((\beta^*)^4)\). With unbalanced presentation, omitted position can shift scores and sensitivities at first order. The symmetry in Corollary~\ref{cor:balanced} is within each judge and comparison. Equal aggregate slot counts alone do not guarantee it. Appendix~\ref{app:A-4}--\ref{app:A-5} give the proofs.

\subsection{Identification}
\label{sec:4-3}

Bias terms are identified only when the design separates their effects from candidate scores. The constraints \(\sum_i s_i=0\) and \(\gamma_{\mathrm{ref}}=1\) fix the overall score level and scale, but other parameter changes can still leave every prediction unchanged. At a finite parameter, full column rank of the predictor Jacobian with respect to the free coordinates \(\theta\) defined in Section~\ref{sec:3-4} is sufficient for local identification. The following proposition shows how family effects can remain confounded with candidate scores.

\statementheading{Proposition}{prop:confounding}{ (Confounding between candidate scores and family effects)} Suppose all candidates and judges belong to two families. Let \(c_i=\mathbf{1}\{\mathrm{fam}(i)=1\}-M^{-1}\sum_j\mathbf{1}\{\mathrm{fam}(j)=1\}\), and let \(\omega_k=1\) for a family-1 judge and \(-1\) for a family-2 judge. Then \(F_t=\omega_{k_t}(c_{a_t}-c_{b_t})\), and for any scalar \(\zeta\) the change

\begin{equation}
\label{eq:8}
s_i' = s_i + \zeta c_i, \qquad \beta_{\mathrm{fam},k}' = \beta_{\mathrm{fam},k} - \zeta\gamma_k\omega_k
\end{equation}

preserves every recorded mean and both constraints. If both candidate families occur, a cross-family target \(q_{ij}\) changes with \(\zeta\), so candidate scores and family effects are not separately identified.

For a concrete case, take one candidate and one unit-sensitivity judge from each family, with position and length effects set to zero. Scores \((0,0)\) with family coefficients \((0,0)\), and scores \((1/2,-1/2)\) with family coefficients \((-1,1)\), both predict a win probability of \(1/2\) for either judge in either display order. In the second setting, each judge\textquotesingle s family effect exactly offsets the candidate score difference. The observations therefore cannot distinguish equal candidate scores from a score difference of one. Correspondingly, the win-probability target \(q_{ij}\) defined in Equation~\eqref{eq:3} equals \(1/2\) in the first setting and \(\sigma(1)\approx0.731\) in the second.

A judge whose family contains no candidate has \(F_t=0\) on all its records. Its predictions therefore move with \(\zeta\) whenever it compares candidates from different families, so these comparisons distinguish the parameter settings in Equation~\eqref{eq:8}. Appendix~\ref{app:A-6} proves Proposition~\ref{prop:confounding}.

\section{Inference}
\label{sec:5}

Judgments of the same prompt can be highly dependent because they share the prompt and its difficulty and may evaluate the same responses. We therefore cluster inference by prompt. Index judgments by \(t=1,\ldots,N\) and prompt clusters by \(g=1,\ldots,G\). Let \(g(t)\) identify the prompt cluster containing judgment \(t\), and write \(\mathcal I_g=\{t:g(t)=g\}\). For any \(\theta\), let \(p_t(\theta)=\sigma\{\eta_t(\theta)\}\) and define

\[
\begin{aligned}
r_t(\theta)&=\nabla_\theta\eta_t(\theta),\\
\psi_t(\theta)&=\{Y_t-p_t(\theta)\}r_t(\theta),\\
\Psi_g(\theta)&=\sum_{t\in\mathcal I_g}\psi_t(\theta).
\end{aligned}
\]

Here \(\psi_t(\theta)\) is the negative gradient of judgment \(t\)\textquotesingle s unpenalized loss contribution, and \(\Psi_g(\theta)\) sums these gradients within prompt \(g\). Write \(\hat p_t=p_t(\hat\theta)\), \(\hat r_t=r_t(\hat\theta)\), \(\widehat\Psi_g=\Psi_g(\hat\theta)\) and \(\overline{\widehat\Psi}=G^{-1}\sum_g\widehat\Psi_g\). The covariance estimate is

\begin{equation}
\label{eq:9}
\begin{aligned}
\hat H
&=\sum_{t=1}^{N}\left\{\hat p_t(1-\hat p_t)\hat r_t\hat r_t^\top-(Y_t-\hat p_t)\nabla_\theta^2\eta_t(\hat\theta)\right\}+\lambda I,\\
\hat V
&=\frac{G}{G-1}\sum_{g=1}^{G}(\widehat\Psi_g-\overline{\widehat\Psi})(\widehat\Psi_g-\overline{\widehat\Psi})^\top,\\
\hat\Sigma
&=\hat H^{-1}\hat V\hat H^{-\top}.
\end{aligned}
\end{equation}

The residual-dependent term in \(\hat H\) arises from the second derivatives of \(\eta_t\), which is nonlinear in the jointly estimated scores and log-sensitivities. In \(\hat V\), we measure variation across the summed gradients \(\widehat\Psi_g\) after subtracting their mean.

\statementheading{Theorem}{thm:cluster}{ (Cluster inference with fixed candidates and judges)} For independent, identically distributed prompt clusters, fixed sets of candidates and judges, and a correctly specified, identified mean model, under the regularity and penalty conditions in Appendix~\ref{app:A-7}, we have

\[
\sqrt{G}(\hat\theta - \theta^*) \xrightarrow{d} N(0,H_{\mathrm{cl}}^{-1}V_{\mathrm{cl}}H_{\mathrm{cl}}^{-\top}) 
\]

where \(H_{\mathrm{cl}} = E[-\nabla_\theta\Psi_g(\theta^*)]\) is the expected curvature per prompt and \(V_{\mathrm{cl}} = E[\Psi_g(\theta^*)\Psi_g(\theta^*)^\top]\) is the covariance matrix of the summed gradient for a prompt. Moreover, \(G\hat\Sigma = G \hat H^{-1}\hat V\hat H^{-\top}\) consistently estimates the limiting covariance \(H_{\mathrm{cl}}^{-1}V_{\mathrm{cl}}H_{\mathrm{cl}}^{-\top}\). Smooth scalar targets with positive limiting variance therefore have asymptotically valid delta-method intervals.

For the win probability \(q_{ij}=\sigma(h_{ij})\) in Equation~\eqref{eq:3}, write

\[
h_{ij}^*=h_{ij}(\theta^*),\qquad
q_{ij}^*=q_{ij}(\theta^*),\qquad
\hat h_{ij}=h_{ij}(\hat\theta),\qquad
\widehat q_{ij}=\sigma(\hat h_{ij}).
\]

We use a first-order Taylor approximation, called the delta method, to estimate the variance of \(\hat h_{ij}\), construct a normal interval on the log-odds scale, and transform its endpoints through \(\sigma\):

\begin{equation}
\label{eq:10}
\begin{aligned}
\widehat{\mathrm{se}}_{ij}^{\,2}
&=
\{\nabla_\theta h_{ij}(\hat\theta)\}^\top
\hat\Sigma
\nabla_\theta h_{ij}(\hat\theta),\\
\mathrm{CI}_{0.95}(q_{ij}^*)
&=
\left[
\sigma\!\left(\hat h_{ij}-1.96\,\widehat{\mathrm{se}}_{ij}\right),
\sigma\!\left(\hat h_{ij}+1.96\,\widehat{\mathrm{se}}_{ij}\right)
\right].
\end{aligned}
\end{equation}

Here \(\widehat{\mathrm{se}}_{ij}\) is the estimated standard error of \(\hat h_{ij}\) on the log-odds scale. Appendix~\ref{app:A-8} derives the target gradient. We use the joint covariance to account for uncertainty in the estimated scores, sensitivities and bias coefficients.

\textbf{Remark.} Mean adjustment and covariance adjustment address different errors. The sandwich covariance estimate in Equation~\eqref{eq:9} accounts for dependence within prompts, but it does not correct the center of an interval fitted under an omitted-position model. Section~\ref{sec:7} illustrates the importance of both adjustments when bias and within-prompt dependence are present.

\section{Experiments}
\label{sec:6}

We evaluate whether the attenuation predicted in Section~\ref{sec:4} appears in real judgments, how much position adjustment improves prediction, and whether it changes which judges or candidates would be selected.

\subsection{Data and protocol}
\label{sec:6-1}

We reanalyze 65,208 judgments released with \cite{xu2026a} for In-House, UltraFeedback, MT-Bench and Chatbot Arena \citep{cui2024,zheng2023,chiang2024}. The four datasets contain 18, 20, 20 and 10 judges, respectively. Prompt clusters are questions on In-House and released prompt identifiers elsewhere. In In-House, the 990 questions correspond one-to-one to the 990 unordered pairs of the 45 candidates. Holding out questions therefore also holds out candidate pairs. Appendix~\ref{app:C} gives sample sizes, preprocessing, prompt grouping, matched-swap construction and position summaries.

\subsection{Does omitted position attenuate judge sensitivity?}
\label{sec:6-2}

We first test whether sensitivity ratios measured on fixed scores follow the attenuation prediction \(1-D^2\) as in Equation~\eqref{eq:5}. We then examine whether attenuation also appears in joint estimation of scores and sensitivities and after swap-and-average. All 18 In-House judges are retained in model fitting and Table~\ref{tab:D2}. The In-House summaries in this subsection use the same 15 judges with finite, positive ratios in the fixed-score experiment.

GLM-4.5-Air has a zero sensitivity ratio, while DeepSeek-R1-Distill-Llama-70B and Mixtral-8x7B have zero adjusted sensitivities and therefore undefined ratios. Their estimates are reported in Table~\ref{tab:D2}. They are omitted from Figure~\ref{fig:1}(c) and the fixed-score log-ratio summaries.

\textbf{Design.} We split the 990 In-House questions into 594 calibration, 198 refit and 198 test questions, keeping all judges, repetitions and display orders of a question together. On the calibration set, we first estimate each judge\textquotesingle s position effect \(\hat\beta_k\) from its exact-swap groups by conditional logistic regression with a Jeffreys correction (Appendix~\ref{app:D-2}). We hold these calibrated position effects fixed and jointly estimate candidate scores and judge sensitivities. We subtract their mean and divide the centered scores by their root mean square to obtain \(\tilde s\).

On the separate refit set, we hold \(\tilde s\) fixed and re-estimate each judge\textquotesingle s sensitivity under two models. One omits position. The other includes the calibrated position effect \(\hat\beta_k\), held fixed:

\begin{equation}
\label{eq:11}
\hat p^{\mathrm N}_t = \sigma\left(\widehat\gamma^{\mathrm N}_{k_t}\tilde d_t\right), \qquad \hat p^{\mathrm P}_t = \sigma\left(\widehat\gamma^{\mathrm P}_{k_t}\tilde d_t + \hat\beta_{k_t}\right), \qquad \tilde d_t = \tilde s_{a_t} - \tilde s_{b_t}
\end{equation}

Superscript \(\mathrm N\) denotes the naive fit without the position term, and \(\mathrm P\) denotes the position-adjusted fit. A bootstrap with 500 replications resamples questions within each fixed split and re-estimates the calibration scores, position effects and refit sensitivities.

\textbf{Sensitivities rise when position is modeled.} Across the 15 judges, 14 sensitivities increase after including the position effect (Table~\ref{tab:D2}), and six paired 95\% intervals for the increase exclude zero. For example, Llama-4-Scout increases from 0.356 to 0.672 (difference 0.316, 95\% CI 0.181--0.531) and gemma-3n-E4B from 0.289 to 0.610 (difference 0.321, 95\% CI 0.107--0.575).

\textbf{Observed attenuation matches predicted \(1-D^2\).} Across the 15 judges, the observed ratio \(\widehat\gamma^{\mathrm N}_k/\widehat\gamma^{\mathrm P}_k\) closely follows \(1-\hat D_k^2\), where \(\hat D_k=\tanh(\hat\beta_k/2)\) (Figure~\ref{fig:1}(c)). The correlation between log observed and log predicted ratios is 0.99. The median absolute log discrepancy is the median across judges of the absolute difference between the log observed and log predicted ratios. Its value is 0.015, and the geometric-mean observed-to-predicted ratio is 1.03 (95\% CI 0.98--1.08).

\textbf{Joint estimation.} We fit J and J+P separately on each full dataset, jointly estimating candidate scores and judge sensitivities in each fit. To compare sensitivities across the resulting score scales, we use the scale-invariant quantity \(\kappa_k = \gamma_k\cdot\mathrm{RMS}(s)\) where \(\mathrm{RMS}(s)=\sqrt{M^{-1}\sum_i s_i^2}\). The ratios \(\widehat\kappa_k^{\mathrm J}/\widehat\kappa_k^{\mathrm{J+P}}\) remain close to the position-based predictions \(1-(\hat D_k^{\mathrm{J+P}})^2\), where \(\hat D_k^{\mathrm{J+P}}=\tanh(\hat\beta^{\mathrm{J+P}}_{\mathrm{pos},k}/2)\) is computed from the J+P fit. The median absolute log discrepancy ranges from 0.004 to 0.058 across the four datasets (Table~\ref{tab:D1}). Thus, the attenuation also appears when candidate scores and judge sensitivities are estimated together.

\textbf{Swap-and-average.} Averaging matched orders on In-House does not fully restore sensitivity. Across the same 15 judges, the scale-invariant ratio \(\widehat\kappa_k^{\mathrm{SwapAvg}}/\widehat\kappa_k^{\mathrm{J+P}}\) has a median of 0.930 (IQR 0.871--0.999), with 11 ratios below one. The ratios are 0.54 for Llama-4-Scout and 0.71 for gemma-3n-E4B. Relative to the position-based prediction \(1-(\hat D^{\mathrm{J+P}}_k)^2\), the median absolute log discrepancy is 0.049 and the geometric-mean observed-to-predicted ratio is 1.02.

\subsection{Held-out prediction}
\label{sec:6-3}

We evaluate whether position adjustment improves prediction on unseen questions and whether length, family and learned judge sensitivities provide additional gains.

Table~\ref{tab:1} compares J, J+P, J+L, J+F, BT+PLF, Davidson+PLF and OSCAR on identical three-fold splits within each dataset. BT+PLF and Davidson+PLF fix judge sensitivities to one. Appendix~\ref{app:E} gives the model specifications and evaluation protocol.

Position provides the largest single-term improvement on every dataset: J+P reduces held-out log loss relative to J by 14--23\%. On In-House, the reduction is 0.1414 (95\% CI 0.1369--0.1453). Length- or family-only adjustment has smaller, dataset-dependent effects (Table~\ref{tab:1}). For example, beyond position adjustment, the gain from adding length and family terms on Chatbot Arena is significant, reducing held-out log loss by 0.0070 (95\% CI 0.0031--0.0095).

Learning judge sensitivities also improves prediction after bias adjustment: OSCAR achieves lower held-out log loss than BT+PLF on all four datasets, with paired 95\% percentile intervals for the improvement excluding zero throughout (Appendix~\ref{app:E-3}).

\begin{table}[!htbp]
\centering
\small
\setlength{\tabcolsep}{4pt}
\renewcommand{\arraystretch}{1.17}
\begin{tabular}{>{\raggedright\arraybackslash}p{\dimexpr 0.160\linewidth-2\tabcolsep\relax}>{\raggedright\arraybackslash}p{\dimexpr 0.107\linewidth-2\tabcolsep\relax}>{\raggedright\arraybackslash}p{\dimexpr 0.110\linewidth-2\tabcolsep\relax}>{\raggedright\arraybackslash}p{\dimexpr 0.110\linewidth-2\tabcolsep\relax}>{\raggedright\arraybackslash}p{\dimexpr 0.110\linewidth-2\tabcolsep\relax}>{\raggedright\arraybackslash}p{\dimexpr 0.120\linewidth-2\tabcolsep\relax}>{\raggedright\arraybackslash}p{\dimexpr 0.180\linewidth-2\tabcolsep\relax}>{\raggedright\arraybackslash}p{\dimexpr 0.103\linewidth-2\tabcolsep\relax}}
\toprule
\textbf{Dataset} & \textbf{J} & \textbf{J+P} & \textbf{J+L} & \textbf{J+F} & \textbf{BT+PLF} & \textbf{Davidson+PLF} & \textbf{OSCAR} \\
\midrule
In-House & 0.6232 & 0.4818 & 0.6223 & 0.6233 & 0.4877 & 0.4874 & 0.4813 \\
UltraFeedback & 0.6089 & 0.5045 & 0.6054 & 0.6087 & 0.5054 & 0.5007 & 0.4998 \\
MT-Bench & 0.5206 & 0.4366 & 0.5187 & 0.5202 & 0.4499 & 0.4416 & 0.4363 \\
Chatbot Arena & 0.6147 & 0.5286 & 0.6086 & 0.6123 & 0.5365 & 0.5313 & 0.5217 \\
\bottomrule
\end{tabular}
\caption{Mean held-out log loss (lower is better). All methods use identical folds within each dataset. Model specifications, fitting settings and paired uncertainty estimates are given in Appendices~\ref{app:B} and~\ref{app:E}.}
\label{tab:1}
\end{table}

\subsection{Judge selection and human agreement}
\label{sec:6-4}

We examine two practical consequences of position adjustment: whether it changes which judges would be excluded by sensitivity-based screening, and whether the resulting candidate rankings better agree with human judgments. We compare judge selection under J and J+P across four datasets and assess human agreement on MT-Bench.

\begin{figure}[!htbp]
\centering
\includegraphics[width=1\linewidth]{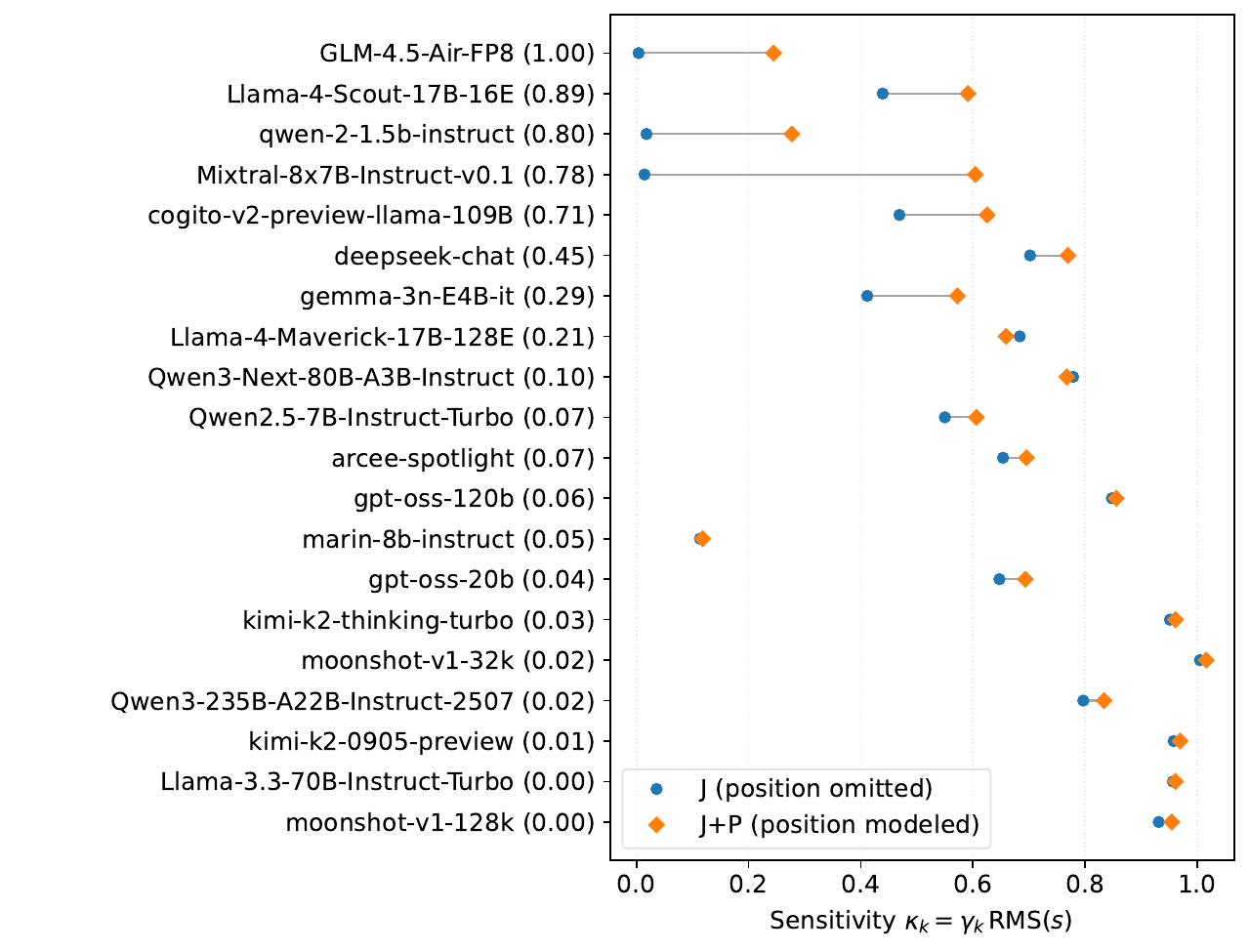}
\caption{Judge sensitivity on UltraFeedback under J (circles) and J+P (diamonds), measured by \(\kappa_k=\gamma_k\,\mathrm{RMS}(s)\). Judges are ordered by the absolute A/B imbalance shown in parentheses. Mixtral-8x7B moves from 19th to 15th of 20 judges after position adjustment, leaving the bottom-three set.}
\label{fig:2}
\end{figure}

\textbf{Judge selection.} Under the full-data J and J+P fits (Appendix~\ref{app:B}), the bottom-three judge sets ranked by \(\kappa_k=\gamma_k\,\mathrm{RMS}(s)\) share 2 of 3 judges on In-House, 2 of 3 on UltraFeedback, 1 of 3 on MT-Bench and 3 of 3 on Chatbot Arena, so position adjustment changes the set on three of the four datasets. On UltraFeedback, Mixtral-8x7B moves from 19th to 15th among 20 judges and leaves the bottom-three set, with \(\widehat\kappa\) increasing from 0.014 to 0.604 (Figure~\ref{fig:2}). A screening rule based on low fitted sensitivity can therefore exclude different judges depending on whether position is modeled. Position adjustment also changes the top-ranked candidate on MT-Bench. J ranks claude-v1 first, whereas J+P and OSCAR rank GPT-4 first, matching the human reference ranking.

\textbf{Human ranking agreement.} We use 1,689 first-turn human votes on exactly matched MT-Bench responses. To construct a human reference ranking, we fit a Bradley--Terry model to these votes (Appendix~\ref{app:F-2}). Candidates are ranked by their fitted human scores. We compare this ordering with rankings from J, J+P and OSCAR, whose candidate scores are estimated using LLM judgments only. Among the six candidates, J+P and OSCAR match the human reference ordering exactly (Spearman 1.00), whereas J reverses the top two candidates, GPT-4 and claude-v1 (Spearman 0.94).

\textbf{Prediction of human votes.} We test whether candidate scores learned from LLM judgments can predict human preferences between responses. For each of J, J+P and OSCAR, we keep the candidate scores fixed at their full-data estimates. For a comparison between candidates \(a\) and \(b\), we predict the human preference probability as

\[
\hat p=\sigma\!\left\{\widehat\alpha_0^H+\widehat\alpha_1^H(\widehat s_a-\widehat s_b)\right\}.
\]

The superscript \(H\) denotes human. Here \(\widehat s_a,\widehat s_b\) are the fixed candidate scores learned from LLM judgments. We estimate the calibration intercept \(\alpha_0^H\) and slope \(\alpha_1^H\) from human votes to translate score differences into preference probabilities \(\widehat p\).

We divide the 80 MT-Bench prompts into five groups. In each of five rounds, we estimate \(\alpha_0^H,\alpha_1^H\) using human votes from four groups and predict votes in the remaining group. We repeat this procedure so that each group is evaluated once. J+P reduces prediction log loss from 0.5687 to 0.5604 relative to J, a reduction of 0.0083 (95\% CI 0.0022--0.0140). Lower log loss indicates better predictions of human votes. Appendix~\ref{app:F} describes the calibration and prompt bootstrap procedures.

\section{Simulations}
\label{sec:7}

The real-data analyses establish changes in fitted sensitivity, prediction and selection, but the true sensitivities and candidate ordering are unknown. We therefore use simulations with known generating parameters to evaluate recovery of judge sensitivities and candidate rankings, estimation of the vector of pairwise win probabilities \(q=(q_{ij})_{i<j}\), and coverage of their 95\% confidence intervals.

We simulate eight candidates and five judges, with every judge evaluating the same displayed response pair once per prompt. The main setting includes position, length and family effects, makes display order depend on verbosity, and induces dependence through shared Gaussian noise within each prompt, with within-prompt latent correlation \(\rho=0.6\) (Appendix~\ref{app:G-1}). We use \(N \in \{1000, 2500, 5000, 10000\}\) judgments and three controls that remove bias, dependence or both, with 100 replications each. Appendix~\ref{app:G} gives the generating parameters and Monte Carlo uncertainty calculations.

\begin{figure}[!htbp]
\centering
\begin{minipage}{1.0\linewidth}\centering
{\small \begin{minipage}[t]{0.49\linewidth}\centering (a) Win-probability RMSE\end{minipage}\hfill\begin{minipage}[t]{0.49\linewidth}\centering (b) Confidence interval coverage\end{minipage}\par}
\smallskip
\includegraphics[width=\linewidth]{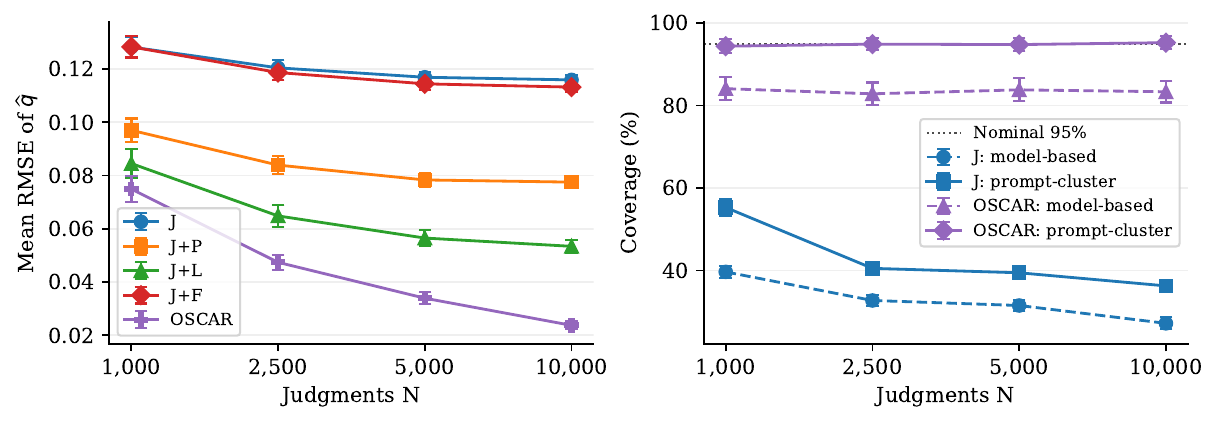}
\end{minipage}
\caption{Estimation of pairwise win probabilities \(q\) in the main simulation setting, with 100 replications per sample size. \textbf{(a)} RMSE across the 28 candidate pairs, computed within each replication and then averaged. \textbf{(b)} Coverage of 95\% confidence intervals for J and OSCAR with model-based and prompt-cluster covariances. Model-based intervals treat judgments as independent. Prompt-cluster intervals account for dependence within prompts. Error bars show 95\% Monte Carlo intervals. The dotted line marks 95\% coverage.}
\label{fig:3}
\end{figure}

\textbf{Point estimates.} At \(N=10{,}000\), OSCAR estimates \(q\) with a root mean squared error (RMSE) of 0.0237, compared with 0.1158 for J and 0.0534 for J+L, the strongest single adjustment in this setting (Figure~\ref{fig:3}(a)). Length adjustment is particularly useful here because the generator links display order to verbosity.

\textbf{Confidence interval coverage.} In the main biased, dependent setting, OSCAR\textquotesingle s prompt-cluster confidence intervals achieve 94.4--95.2\% average coverage across sample sizes (Figure~\ref{fig:3}(b)). Using the same point estimates but computing model-based intervals as if individual judgments were independent gives only 82.8--84.1\% coverage. Conversely, cluster covariance does not correct a mean model with omitted terms: coverage of J\textquotesingle s prompt-cluster intervals falls from 55.3\% to 36.3\% as the sample size increases. Both the mean specification and the dependence adjustment matter.

The controls (Table~\ref{tab:G2}) separate mean misspecification from within-prompt dependence. With bias removed but within-prompt dependence retained, J\textquotesingle s prompt-cluster intervals achieve 95.6\% coverage. With dependence removed but bias retained, OSCAR\textquotesingle s model-based intervals achieve 95.0\% coverage. Therefore, achieving coverage close to 95\% requires both modeling the bias effects and accounting for dependence among judgments of the same prompt.

\begin{figure}[!htbp]
\centering
\begin{minipage}{1.0\linewidth}\centering
{\small \begin{minipage}[t]{0.49\linewidth}\centering (a) Recovery of judge sensitivities\end{minipage}\hfill\begin{minipage}[t]{0.49\linewidth}\centering (b) Recovery of candidate rankings\end{minipage}\par}
\smallskip
\includegraphics[width=\linewidth]{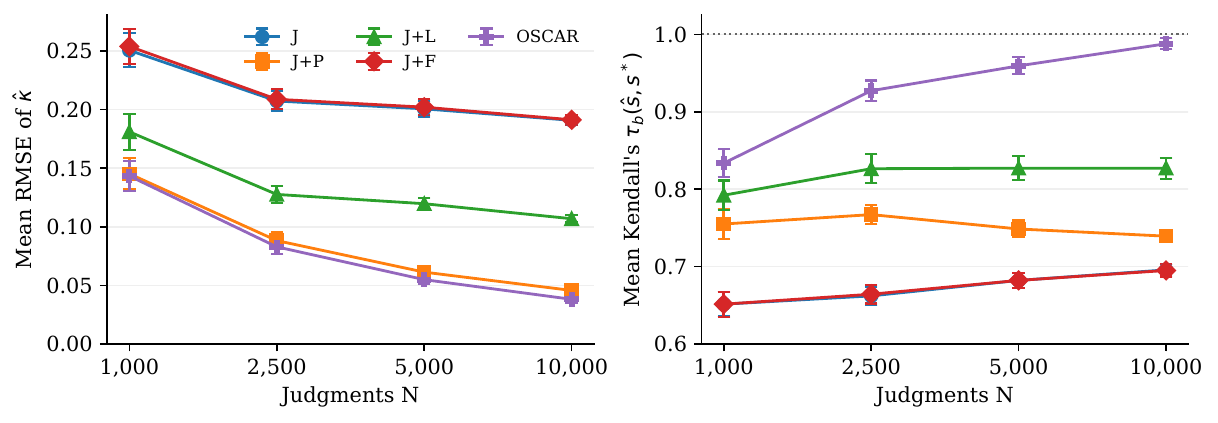}
\end{minipage}
\caption{Recovery of judge sensitivities and candidate rankings in the main biased, dependent setting. \textbf{(a)} Mean replication-level RMSE of \(\widehat\kappa\) against the generating values \(\kappa_k^*=\gamma_k^*\mathrm{RMS}(s^*)\) (lower is better). \textbf{(b)} Mean Kendall\textquotesingle s \(\tau_b(\widehat s,s^*)\) between the estimated candidate scores and the true generating scores (higher is better). Error bars show 95\% Monte Carlo intervals across 100 replications. The dotted line in panel (b) marks perfect agreement with the true candidate ranking, \(\tau_b=1\).}
\label{fig:4}
\end{figure}

\textbf{Judge sensitivity recovery.} We compare the estimated scale-invariant sensitivities \(\widehat\kappa_k=\widehat\gamma_k\,\mathrm{RMS}(\widehat s)\) with their generating values \(\kappa_k^*=\gamma_k^*\,\mathrm{RMS}(s^*)\), computing an RMSE across judges within each replication and then averaging over replications. OSCAR\textquotesingle s mean RMSE decreases from 0.1434 at \(N=1000\) to 0.0381 at \(N=10{,}000\) (Figure~\ref{fig:4}(a)). At \(N=10{,}000\), the RMSE is 0.1908 for J and 0.0457 for J+P, the best single-term adjustment for sensitivity recovery. OSCAR\textquotesingle s paired RMSE difference from J+P is significant (−0.0076, 95\% Monte Carlo interval −0.0103 to −0.0048).

\textbf{Candidate ranking recovery.} OSCAR\textquotesingle s mean Kendall\textquotesingle s rank correlation \(\tau_b\) (1 for identical orderings) with the true candidate ordering increases from 0.834 at \(N=1000\) to 0.988 at \(N=10{,}000\) (Figure~\ref{fig:4}(b)). At \(N=10{,}000\), OSCAR improves mean Kendall\textquotesingle s \(\tau_b\) over J+L, the best single-term adjustment for ranking recovery, by 0.161 (95\% Monte Carlo interval 0.144--0.177). OSCAR selects the true best candidate in 96 of 100 replications, compared with 12 for J+L.

\section{Conclusion}
\label{sec:8}

Omitting position can attenuate fitted judge sensitivity even under balanced presentation or swap-and-average. OSCAR models judge-specific sensitivity together with position, length and family effects and provides prompt-cluster confidence intervals that account for dependence among judgments of the same prompt. Across four datasets, position adjustment improves held-out prediction and changes the bottom-three judge set on three of them. On MT-Bench, it changes the top-ranked candidate from claude-v1 to GPT-4, and the adjusted candidate ranking matches the human reference ranking. Simulations show improved sensitivity and ranking recovery and confidence interval coverage close to 95\%.

\FloatBarrier

\bibliographystyle{plainnat}
\bibliography{references}

\clearpage\appendix
\numberwithin{equation}{section}
\renewcommand{\theequation}{\thesection\arabic{equation}}
\numberwithin{table}{section}
\renewcommand{\thetable}{\thesection\arabic{table}}
\titleformat{\section}{\large\bfseries}{Appendix \thesection}{0.7em}{}

\section{Proofs and derivations}
\label{app:A}

\subsection{A global attenuation identity}
\label{app:A-1}

We show that averaging the two display orders attenuates sensitivity at every quality gap when \(\beta^*\ne0\).

Fix one judge with true sensitivity \(\gamma^*>0\) and position intercept \(\beta^*\), and consider a response whose quality gap over the other response is \(x\). Let \(o=+1\) when the response is shown first and \(o=-1\) when it is shown second, with the two orders equally likely. Write \(\sigma_o(x)=\sigma(\gamma^*x+o\beta^*)\) and \(m(x)=E_o[\sigma_o(x)]\), where \(E_o\) averages over these two orders. The corresponding variance is

\[
\mathrm{Var}_o\{\sigma_o(x)\}
=\tfrac14\{\sigma_{+1}(x)-\sigma_{-1}(x)\}^2.
\]

\lemmaheading{lem:A1}{} For every \(x\in\mathbb R\),

\begin{equation}
\label{eq:A1}
0<\frac{d\,\mathrm{logit}\,m(x)}{dx}
=\gamma^*\left[
1-\frac{\mathrm{Var}_o\{\sigma_o(x)\}}
{m(x)\{1-m(x)\}}
\right]
\leq\gamma^*.
\end{equation}

The upper inequality is strict whenever \(\beta^*\neq0\).

\emph{Proof.} The chain rule gives

\[
\frac{d\,\mathrm{logit}\,m(x)}{dx}
=\frac{m'(x)}{m(x)\{1-m(x)\}}.
\]

Since \(\sigma'=\sigma(1-\sigma)\) and

\[
E_o[\sigma_o(x)^2]
=\mathrm{Var}_o\{\sigma_o(x)\}+m(x)^2,
\]

we have

\[
\begin{aligned}
m'(x)
&=\gamma^*E_o[\sigma_o(x)\{1-\sigma_o(x)\}]\\
&=\gamma^*\{m(x)-E_o[\sigma_o(x)^2]\}\\
&=\gamma^*\left[m(x)\{1-m(x)\}-\mathrm{Var}_o\{\sigma_o(x)\}\right].
\end{aligned}
\]

Substituting into the chain-rule identity gives the equality in \eqref{eq:A1}. Since \(\gamma^*>0\) and \(0<\sigma_o(x)<1\), we have \(m'(x)=\gamma^*E_o[\sigma_o(x)\{1-\sigma_o(x)\}]>0\). Together with \(0<m(x)<1\), this gives the positive lower bound in \eqref{eq:A1}. If \(\beta^*\neq0\), strict monotonicity of \(\sigma\) implies that \(\sigma(\gamma^*x+\beta^*)\) and \(\sigma(\gamma^*x-\beta^*)\) differ at every finite \(x\). Their variance is therefore positive, giving the strict inequality. \(\square\)

\subsection{Derivation of Equation~\eqref{eq:5}}
\label{app:A-2}

We evaluate Lemma~\ref{lem:A1} at zero quality gap to obtain the attenuation factor in Equation~\eqref{eq:5}. Since \(\sigma(-y)=1-\sigma(y)\),

\[
\begin{aligned}
m(0)&=\tfrac12,\\
\sigma(\beta^*)-\sigma(-\beta^*)
&=\frac{e^{\beta^*}-1}{e^{\beta^*}+1}
=\tanh(\beta^*/2)=D^*.
\end{aligned}
\]

The variance formula in Appendix~\ref{app:A-1} therefore gives

\[
\mathrm{Var}_o\{\sigma_o(0)\}=\frac{(D^*)^2}{4}.
\]

Substituting these values into \eqref{eq:A1},

\[
\left.\frac{d\,\mathrm{logit}\,m(x)}{dx}\right|_{x=0}
=\gamma^*\left\{1-\frac{(D^*)^2/4}{1/4}\right\}=\gamma^*\{1-(D^*)^2\}=\gamma^*\mathrm{sech}^2(\beta^*/2).
\]

Since \(1-\tanh^2(y)=\mathrm{sech}^2(y)\), this proves Equation~\eqref{eq:5}. \(\square\)

\subsection{Population fits under balanced and unbalanced order}
\label{app:A-3}

We first prove attenuation of the population-fitted sensitivity under balanced order and derive the approximation \(\gamma^*\{1-(D^*)^2\}\) when quality gaps are near zero. We then give the calculations for Figure~\ref{fig:1}(b) and the Introduction\textquotesingle s 40:60 example.

\lemmaheading{lem:A2}{ (Attenuation of the fitted sensitivity)} Fix quality gaps \(x_1,\ldots,x_{n_x}\), with at least one nonzero gap. Let \(m(x)\) be the balanced-order mean generated by \(\gamma^*>0\) and \(\beta^*\ne0\), as defined in Appendix~\ref{app:A-1}. The omitted-position population loss

\[
\mathcal L(\gamma)
=\sum_{j=1}^{n_x}\left[\log(1+e^{\gamma x_j})-m(x_j)\gamma x_j\right]
\]

is the summed expected log loss of the predictions \(\sigma(\gamma x_j)\) when the response with gap \(x_j\) has mean preference score \(m(x_j)\). It has a unique minimizer over \(\gamma\geq0\), denoted by \(\gamma^\dagger\), and it satisfies \(0<\gamma^\dagger<\gamma^*\).

\emph{Proof.} By Lemma~\ref{lem:A1}, \(m\) is strictly increasing. Since \(m(0)=1/2\) (Appendix~\ref{app:A-2}), \(m(x)>1/2\) for \(x>0\) and \(m(x)<1/2\) for \(x<0\). Also, \(\mathrm{logit}\,m(0)=0\). Integrating the strict slope bound in Lemma~\ref{lem:A1} from zero to \(x\) gives

\[
\begin{cases}
0<\mathrm{logit}\,m(x)<\gamma^*x, & x>0,\\
\gamma^*x<\mathrm{logit}\,m(x)<0, & x<0.
\end{cases}
\]

Applying the increasing function \(\sigma\) shows that \(m(x)\) lies strictly between \(1/2\) and \(\sigma(\gamma^*x)\) for every nonzero \(x\).

The first two derivatives of the loss are

\[
\begin{aligned}
\mathcal L'(\gamma)
&=\sum_{j=1}^{n_x}x_j\{\sigma(\gamma x_j)-m(x_j)\},\\
\mathcal L''(\gamma)
&=\sum_{j=1}^{n_x}x_j^2\sigma(\gamma x_j)\{1-\sigma(\gamma x_j)\}>0.
\end{aligned}
\]

At \(\gamma=0\), a positive gap has \(m(x_j)>1/2\) and a negative gap has \(m(x_j)<1/2\), so every nonzero-gap term in \(\mathcal L'(0)\) is negative. At \(\gamma=\gamma^*\), the difference \(\sigma(\gamma^*x_j)-m(x_j)\) has the same sign as \(x_j\), so every nonzero-gap term is positive. Zero gaps contribute zero. Consequently,

\[
\mathcal L'(0)<0,
\qquad
\mathcal L'(\gamma^*)>0.
\]

The continuous, strictly increasing derivative has exactly one zero between \(0\) and \(\gamma^*\). Strict convexity makes it the unique global minimizer. \(\square\)

\lemmaheading{lem:A3}{ (Near-zero approximation)} Fix baseline gaps \(x_j^{\mathrm{base}}\) with \(\sum_j(x_j^{\mathrm{base}})^2>0\), and set \(x_j(\delta)=\delta x_j^{\mathrm{base}}\) for \(\delta>0\). Under the conditions of Lemma~\ref{lem:A2}, let \(\gamma^\dagger(\delta)\) minimize the resulting loss \(\mathcal L_\delta\). As \(\delta\downarrow0\),

\[
\gamma^\dagger(\delta)
=\gamma^*\{1-(D^*)^2\}+O(\delta^2).
\]

\emph{Proof.} The logistic expansion at zero is

\[
\sigma(y)=\tfrac12+\tfrac14y+O(y^3).
\]

For the balanced mean, \(m(-x)=1-m(x)\) gives \(m''(0)=0\). Appendix~\ref{app:A-2} and the chain rule give \(m'(0)=\gamma^*\{1-(D^*)^2\}/4\), so

\[
\begin{aligned}
m(x)
&=m(0)+m'(0)x+\tfrac12m''(0)x^2+O(x^3)\\
&=\tfrac12+\tfrac14\gamma^*\{1-(D^*)^2\}x+O(x^3).
\end{aligned}
\]

Lemma~\ref{lem:A2} gives \(0<\gamma^\dagger(\delta)<\gamma^*\) for every \(\delta>0\). Because the baseline gaps \(x_j^{\mathrm{base}}\) are fixed and finite in number, this bound makes the remainder in \(\sigma\{\gamma^\dagger(\delta)\delta x_j^{\mathrm{base}}\}\) uniformly \(O(\delta^3)\). Substituting both expansions into the first-order condition yields

\[
\begin{aligned}
0
&=\mathcal L_\delta'\{\gamma^\dagger(\delta)\}\\
&=\sum_j x_j(\delta)
\left[\sigma\{\gamma^\dagger(\delta)x_j(\delta)\}-m\{x_j(\delta)\}\right]\\
&=\frac{\delta^2}{4}\sum_j(x_j^{\mathrm{base}})^2
\left[\gamma^\dagger(\delta)-\gamma^*\{1-(D^*)^2\}\right]+O(\delta^4).
\end{aligned}
\]

Dividing by \(\delta^2\sum_j(x_j^{\mathrm{base}})^2/4\) proves the expansion. \(\square\)

\textbf{Calculation for Figure~\ref{fig:1}(b).} Set the true sensitivity to \(\gamma^*=1\) and use 401 equally weighted and equally spaced quality gaps on \([-1,1]\). For each position effect \(\beta^*\), we compute the omitted-position population optimum \(\gamma^{\mathrm N,\dagger}(\beta^*)\) by solving the first-order condition for Equation~\eqref{eq:6}. With the position term, the fitted model is \(\mathrm{logit}\,p=\gamma x+\beta o\), where \(o=+1\) if the response is shown first and \(o=-1\) otherwise. Write \(\tilde x_{jo}=(x_j,o)^\top\), and let \(\pi_{jo}>0\) be the weight assigned to gap \(x_j\) and order \(o\). The position-adjusted population loss is

\[
\mathcal L^{\mathrm P}(\gamma,\beta)
=\sum_{j,o}\pi_{jo}\bigl[\log(1+e^{\gamma x_j+\beta o}) -\sigma(\gamma^*x_j+\beta^*o)(\gamma x_j+\beta o)\bigr].
\]

Its Hessian is

\[
\nabla^2\mathcal L^{\mathrm P}(\gamma,\beta)
=\sum_{j,o}\pi_{jo}\,\sigma'(\gamma x_j+\beta o)\,\tilde x_{jo}\tilde x_{jo}^\top.
\]

For any \(h=(h_1,h_2)^\top\),

\[
h^\top\nabla^2\mathcal L^{\mathrm P}(\gamma,\beta)h
=\sum_{j,o}\pi_{jo}\,\sigma'(\gamma x_j+\beta o)(h_1x_j+h_2o)^2.
\]

All weights are positive at finite parameters. Choose a nonzero gap \(x_j\). Since both orders occur at this gap, the quadratic form can vanish only if

\[
h_1x_j+h_2=0,\qquad h_1x_j-h_2=0,
\]

which implies \(h_1=h_2=0\). The Hessian is therefore positive definite, and the loss is strictly convex.

Its first-order conditions hold at \((\gamma,\beta)=(\gamma^*,\beta^*)=(1,\beta^*)\), where the model probabilities equal the true ones. Hence \((1,\beta^*)\) is the unique minimizer and the position-adjusted population optimum satisfies \(\gamma^{\mathrm P,\dagger}(\beta^*)=1\).

\textbf{The Introduction\textquotesingle s 40:60 example.} Set \(\gamma^*=1\), \(\beta^*=1\), and use 200 equally weighted gaps \(x_j=j/200\). The higher-quality response is first with probability \(0.4\). Its order-averaged mean and omitted-position fit are

\[
\begin{aligned}
m_{0.4}(x)&=0.4\sigma(x+1)+0.6\sigma(x-1),\\
\gamma_{0.4}^\dagger&=\arg\min_{\gamma\geq0}\;
\frac1{200}\sum_{j=1}^{200}
\left[\log(1+e^{\gamma x_j})-m_{0.4}(x_j)\gamma x_j\right]
\approx0.528.
\end{aligned}
\]

The calculation uses expected loss without sampling verdicts. The position-adjusted model \(\sigma(\gamma x+\beta o)\), with \(o=+1\) or \(-1\) for the response\textquotesingle s position, has population minimizer \((\gamma,\beta)=(1,1)\) by the same strict-convexity argument above.

\subsection{Proof of Theorem~\ref{thm:omitted-shifts}}
\label{app:A-4}

For the J model under the constraints in Section~\ref{sec:3-3} with the fixed finite candidates and judges of Section~\ref{sec:3}, we first establish identification and continuity of the inverse from predictors to parameters. We then show that all global minimizers must lie near the true parameter when the omitted effects are small, and derive the unique minimizer\textquotesingle s expansion.

Write \(\mu_\varepsilon=\sigma\{\eta^{\mathrm J}(\xi^*)+z^\top\varepsilon\}\) and \(r(\xi)=\nabla_\xi\eta^{\mathrm J}(\xi)\). The population loss and its gradient are

\[
\begin{aligned}
\mathcal L(\xi,\varepsilon)
&=E\left[\log\left(1+e^{\eta^{\mathrm J}(\xi)}\right)-\mu_\varepsilon\eta^{\mathrm J}(\xi)\right],\\
\nabla_\xi\mathcal L(\xi,\varepsilon)
&=E\left[\left\{\sigma(\eta^{\mathrm J}(\xi))-\mu_\varepsilon\right\}r(\xi)\right].
\end{aligned}
\]

Differentiating the gradient under the expectation gives the two Hessian terms

\[
\nabla_\xi^2\mathcal L(\xi,\varepsilon)=E\left[\sigma'(\eta^{\mathrm J}(\xi))r(\xi)r(\xi)^\top\right]
+E\left[\left\{\sigma(\eta^{\mathrm J}(\xi))-\mu_\varepsilon\right\}
\nabla_\xi^2\eta^{\mathrm J}(\xi)\right].
\]

At \((\xi^*,0)\), the model mean equals \(\mu_0=\sigma\{\eta^{\mathrm J}(\xi^*)\}\), so the second term vanishes and \(\nabla_\xi\mathcal L(\xi^*,0)=0\).

Since

\[
\frac{\partial\mu_\varepsilon}{\partial\varepsilon^\top}
=\sigma'\{\eta^{\mathrm J}(\xi^*)+z^\top\varepsilon\}z^\top,
\]

differentiating the loss gradient under the expectation gives

\[
\partial_\varepsilon\nabla_\xi\mathcal L(\xi,\varepsilon)
=-E\left[\sigma'\{\eta^{\mathrm J}(\xi^*)+z^\top\varepsilon\}r(\xi)z^\top\right].
\]

At \((\xi^*,0)\), we therefore have

\[
\nabla_\xi^2\mathcal L(\xi^*,0)=H^*,
\qquad
\partial_\varepsilon\nabla_\xi\mathcal L(\xi^*,0)
=-E[w^*r^*z^\top]=-B^*.
\]

\textbf{Step 1. Identification and continuity of the inverse.} An ordered comparison cell \(j\) consists of all judgments with the same candidates in slots A and B and the same judge, denoted by \((a_j,b_j,k_j)\). Expectations weight these finitely many cells by their population frequencies \(\pi_j>0\) among judgments.

\textbf{(i) Identification.} In the parametrization of Section~\ref{sec:3-4}, \(s=Cu\), \(\gamma_k=e^{v_k}\) for non-reference judges and \(\gamma_{\mathrm{ref}}=1\). Put \(d_j=(C_{a_j}-C_{b_j})^\top\) and

\[
\eta_j^{\mathrm J}(\xi)=\gamma_{k_j}d_j^\top u,
\qquad
\eta_j^{\mathrm J,*}=\eta_j^{\mathrm J}(\xi^*).
\]

A predictor vector \(\eta=(\eta_j)_j\) is feasible if \(\eta=\eta^{\mathrm J}(\xi)\) for some \(\xi=(u,v)\) in this parametrization. Let \(\vartheta=(u,\gamma_{-\mathrm{ref}}^{-1})\), where the inverse is coordinatewise. The inverse-sensitivity coordinates make the equations linear once the predictor vector is fixed. For a feasible \(\eta\), dividing \(\eta_j=\gamma_{k_j}d_j^\top u\) by \(\gamma_{k_j}\) gives \(d_j^\top u=\eta_j\) for reference-judge cells, where \(\gamma_{\mathrm{ref}}=1\), and \(d_j^\top u-\eta_j\gamma_{k_j}^{-1}=0\) for the other cells. Stacking these equations gives the linear system \(X(\eta)\vartheta=y(\eta)\), with

\[
X_j(\eta)=\bigl(d_j^\top,-\eta_j e_{k_j}^\top\bigr),
\qquad
y_j(\eta)=\eta_j\mathbf 1\{k_j=\mathrm{ref}\}.
\]

Here \(e_k\) is the coordinate vector for a non-reference judge and \(e_{\mathrm{ref}}=0\). Let \(R^*\) be the predictor Jacobian at \(\xi^*\), whose \(j\)th row is \((\nabla_\xi\eta_j^{\mathrm J}(\xi^*))^\top=(\gamma^*_{k_j}d_j^\top,\eta_j^{\mathrm J,*}e_{k_j}^\top)\) (using \(\partial\gamma_k/\partial v_k=\gamma_k\) for \(k\ne\mathrm{ref}\)). Then

\[
H^*=(R^*)^\top W^*R^*,
\qquad
W^*=\mathrm{diag}_j\!\left\{\pi_j\sigma'(\eta_j^{\mathrm J,*})\right\}.
\]

All diagonal entries of \(W^*\) are positive, so \(H^*\succ0\) implies that \(R^*\) has full column rank. Define the invertible diagonal matrices

\[
\Gamma_1=\mathrm{diag}_j\!\left\{\frac1{\gamma^*_{k_j}}\right\},
\qquad
\Gamma_2=\mathrm{diag}\!\left(I_{M-1},-\mathrm{diag}_{k\ne\mathrm{ref}}\{\gamma_k^*\}\right).
\]

Dividing row \(j\) of \(R^*\) by \(\gamma^*_{k_j}\) and multiplying each non-reference \(v_k\) column by \(-\gamma_k^*\) gives

\begin{equation}
\label{eq:A2}
X(\eta^{\mathrm J,*})=\Gamma_1R^*\Gamma_2.
\end{equation}

Hence \(X(\eta^{\mathrm J,*})\) also has full column rank. Its linear system uniquely determines \(\vartheta^*\), and therefore \(\xi^*\), so \(\eta^{\mathrm J}(\xi)=\eta^{\mathrm J,*}\) implies \(\xi=\xi^*\).

Global identification here follows from the bilinear dependence of \(\eta^{\mathrm J}\) on scores and sensitivities and the full column rank of \(X(\eta^{\mathrm J,*})\).

\textbf{(ii) Continuity of the inverse.} The matrix \(X(\eta)\) is affine in \(\eta\) and hence continuous. The condition \(\det\{X(\eta)^\top X(\eta)\}>0\) is open, so the full column rank in \eqref{eq:A2} persists near \(\eta^{\mathrm J,*}\). For every feasible \(\eta\) in that neighborhood,

\[
\vartheta=\{X(\eta)^\top X(\eta)\}^{-1}X(\eta)^\top y(\eta).
\]

The right-hand side is continuous in \(\eta\). Its inverse-sensitivity coordinates converge to the positive values at \(\xi^*\), and taking their negative logarithms recovers \(v\) continuously. Thus, for any parameter sequence \(\xi_n=(u_n,v_n)\) in this parametrization,

\begin{equation}
\label{eq:A3}
\eta^{\mathrm J}(\xi_n)\longrightarrow\eta^{\mathrm J,*}
\quad\Longrightarrow\quad
\xi_n\longrightarrow\xi^*.
\end{equation}

\textbf{Step 2. Parameters whose loss is at most \(\mathcal L(\xi^*,\varepsilon)\) lie near \(\xi^*\).} Define \(\bar\mu_j(\varepsilon)=E[\mu_\varepsilon\mid j]\). Because \(\eta^{\mathrm J}\) is constant within each cell, the population loss is \(\mathcal L(\xi,\varepsilon)=\mathcal L_{\mathrm{pred}}(\eta^{\mathrm J}(\xi),\varepsilon)\), where

\[
\mathcal L_{\mathrm{pred}}(\eta,\varepsilon)
=\sum_j\pi_j\left\{\log(1+e^{\eta_j})-\bar\mu_j(\varepsilon)\eta_j\right\}.
\]

Bounded convergence gives \(\bar\mu_j(\varepsilon)\to\sigma(\eta_j^{\mathrm J,*})\) as \(\varepsilon\to0\). There are finitely many cells and \(\xi^*\) is finite, so some \(p_{\min}>0\) satisfies \(p_{\min}\leq\bar\mu_j(\varepsilon)\leq1-p_{\min}\) for every cell and all sufficiently small \(\varepsilon\). For \(\mu\in[p_{\min},1-p_{\min}]\),

\[
\begin{aligned}
x\geq0:\quad &\log(1+e^x)-\mu x\geq(1-\mu)x\geq p_{\min}|x|,\\
x<0:\quad &\log(1+e^x)-\mu x\geq-\mu x\geq p_{\min}|x|.
\end{aligned}
\]

It follows that

\[
\mathcal L(\xi,\varepsilon)\geq p_{\min}\sum_j\pi_j|\eta_j^{\mathrm J}(\xi)|.
\]

Consider any \(\varepsilon_n\to0\) and any parameter sequence with \(\mathcal L(\xi_n,\varepsilon_n)\leq\mathcal L(\xi^*,\varepsilon_n)\). The right-hand side is bounded. Choose \(\mathcal L_{\mathrm{bd}}<\infty\) with \(\mathcal L(\xi^*,\varepsilon_n)\leq \mathcal L_{\mathrm{bd}}\) for all \(n\), and put \(\pi_{\min}=\min_j\pi_j>0\). The preceding loss bound gives, for all sufficiently large \(n\),

\[
\max_j|\eta_j^{\mathrm J}(\xi_n)|
\leq\frac{\mathcal L_{\mathrm{bd}}}{p_{\min}\pi_{\min}}.
\]

Thus the predictor vectors \(\eta^{\mathrm J}(\xi_n)\) are bounded. Because \(\mathcal L_{\mathrm{pred}}\) is continuous in \((\eta,\varepsilon)\), every convergent subsequence has a limit \(\eta^{\mathrm{lim}}\) satisfying \(\mathcal L_{\mathrm{pred}}(\eta^{\mathrm{lim}},0)\leq\mathcal L_{\mathrm{pred}}(\eta^{\mathrm J,*},0)\). On the unrestricted predictor space, \(\mathcal L_{\mathrm{pred}}(\cdot,0)\) is strictly convex, with

\[
\frac{\partial\mathcal L_{\mathrm{pred}}(\eta,0)}{\partial\eta_j}
=\pi_j\{\sigma(\eta_j)-\sigma(\eta_j^{\mathrm J,*})\}.
\]

It is therefore uniquely minimized at \(\eta^{\mathrm J,*}\). Every convergent subsequence of the bounded predictor sequence has this same limit, so the entire sequence converges to \(\eta^{\mathrm J,*}\). By \eqref{eq:A3}, \(\xi_n\to\xi^*\).

Consequently, for every neighborhood \(\mathcal N\) of \(\xi^*\) and all sufficiently small \(\varepsilon\),

\begin{equation}
\label{eq:A4}
\{\xi:\mathcal L(\xi,\varepsilon)\leq\mathcal L(\xi^*,\varepsilon)\}\subset\mathcal N.
\end{equation}

If \eqref{eq:A4} failed, there would be \(\varepsilon_n\to0\) and \(\xi_n\notin\mathcal N\) with \(\mathcal L(\xi_n,\varepsilon_n)\leq\mathcal L(\xi^*,\varepsilon_n)\). The argument above gives \(\xi_n\to\xi^*\), a contradiction.

\textbf{Step 3. Global minimization and expansion.} Since \(\nabla_\xi^2\mathcal L(\xi^*,0)=H^*\succ0\) and the Hessian is continuous, choose a small closed ball \(\mathcal B\) centered at \(\xi^*\) on which the Hessian remains positive definite for all sufficiently small \(\varepsilon\). The continuous loss attains a minimum on \(\mathcal B\), with value at most \(\mathcal L(\xi^*,\varepsilon)\). Applying \eqref{eq:A4} with \(\mathcal N=\mathrm{int}(\mathcal B)\) shows that this minimum lies in the interior and every parameter outside \(\mathcal B\) has larger loss. It is therefore a global minimizer over the full parameter space under these constraints. Any other global minimizer also has loss at most \(\mathcal L(\xi^*,\varepsilon)\), so by \eqref{eq:A4} it lies in \(\mathcal B\), where the loss is strictly convex. The global minimizer is therefore unique.

The implicit function theorem, applied to \(\nabla_\xi\mathcal L\) with \(\nabla_\xi^2\mathcal L(\xi^*,0)=H^*\), gives a unique twice continuously differentiable stationary branch near \((\xi^*,0)\). The global minimizer lies in \(\mathrm{int}(\mathcal B)\) and is therefore stationary. By \eqref{eq:A4}, it also lies in the implicit function theorem\textquotesingle s uniqueness neighborhood for all sufficiently small \(\varepsilon\), so it equals that branch. Its derivative at zero is

\[
\begin{aligned}
\left.\frac{\partial\xi^\dagger(\varepsilon)}{\partial\varepsilon^\top}\right|_{\varepsilon=0}
&=-\{\nabla_\xi^2\mathcal L(\xi^*,0)\}^{-1}
\partial_\varepsilon\nabla_\xi\mathcal L(\xi^*,0)\\
&=(H^*)^{-1}B^*.
\end{aligned}
\]

Taylor\textquotesingle s theorem yields

\[
\xi^\dagger(\varepsilon)-\xi^*
=(H^*)^{-1}B^*\varepsilon+O(\lVert\varepsilon\rVert^2),
\]

which is Equation~\eqref{eq:7}. \(\square\)

\subsection{Proof of Corollary~\ref{cor:balanced}}
\label{app:A-5}

We show that balanced presentation removes the first-order parameter shift caused by omitted position effects. Let \(T\) map a record \((a,b,L,F,k)\) to \((b,a,-L,-F,k)\). Invariance of the design distribution under \(T\) means \(E[f\circ T]=E[f]\) for every integrable function \(f\) of the record.

For every \(\xi\), the J predictor satisfies \(\eta^{\mathrm J}(\xi)\circ T=-\eta^{\mathrm J}(\xi)\). Differentiating this identity with respect to \(\xi\) gives \(\{\nabla_\xi\eta^{\mathrm J}(\xi)\}\circ T=-\nabla_\xi\eta^{\mathrm J}(\xi)\), so \(r^*=\nabla_\xi\eta^{\mathrm J}(\xi^*)\) changes sign under \(T\). The weight \(w^*=\sigma'\{\eta^{\mathrm J}(\xi^*)\}\) is unchanged because \(\sigma'(-y)=\sigma'(y)\). The position column for judge \(k'\) is \(z_{\mathrm{pos},k'}=\mathbf 1\{k=k'\}\), which \(T\) leaves unchanged. Invariance therefore gives

\[
E[w^*r^*z_{\mathrm{pos},k'}]
=E[(w^*r^*z_{\mathrm{pos},k'})\circ T]
=-E[w^*r^*z_{\mathrm{pos},k'}].
\]

Thus every position column of \(B^*\) is zero. When position is the only omitted effect, \(\varepsilon\) has only position coordinates, so \(B^*\varepsilon=0\) and Equation~\eqref{eq:7} gives \(\xi^\dagger(\varepsilon)-\xi^*=O(\lVert\varepsilon\rVert^2)\). \(\square\)

\subsection{Proof of Proposition~\ref{prop:confounding}}
\label{app:A-6}

We show that the family contrast can be absorbed into candidate scores while preserving every recorded mean and both constraints. The centering term in \(c_i\) cancels in differences. For a family-1 judge,

\[
F_t=\mathbf 1\{\mathrm{fam}(a_t)=1\}-\mathbf 1\{\mathrm{fam}(b_t)=1\}
=c_{a_t}-c_{b_t}.
\]

There are only two candidate families, so \(\mathbf 1\{\mathrm{fam}(i)=2\}=1-\mathbf 1\{\mathrm{fam}(i)=1\}\). For a family-2 judge, this gives

\[
F_t
=[1-\mathbf 1\{\mathrm{fam}(a_t)=1\}]-[1-\mathbf 1\{\mathrm{fam}(b_t)=1\}]
=-(c_{a_t}-c_{b_t}).
\]

Hence \(F_t=\omega_{k_t}(c_{a_t}-c_{b_t})\), with \(\omega_k^2=1\) for every judge. Under Equation~\eqref{eq:8}, the score term changes by \(\zeta\gamma_{k_t}(c_{a_t}-c_{b_t})\) and the family term by \(-\zeta\gamma_{k_t}\omega_{k_t}F_t\), so

\[
\begin{aligned}
\eta'_t-\eta_t
&=\zeta\gamma_{k_t}\left[(c_{a_t}-c_{b_t})-\omega_{k_t}F_t\right]\\
&=\zeta\gamma_{k_t}(1-\omega_{k_t}^2)(c_{a_t}-c_{b_t})=0.
\end{aligned}
\]

Both constraints remain satisfied because \(\sum_i c_i=0\) and no sensitivity changes. When both candidate families occur, there is a pair \(i,j\) from different families. For this pair, \(c_i-c_j=\pm1\). Because \(\bar\gamma\) does not change,

\[
h'_{ij}=h_{ij}+\bar\gamma\zeta(c_i-c_j).
\]

Since \(\bar\gamma>0\) and \(\sigma\) is strictly increasing, the cross-family target \(q_{ij}=\sigma(h_{ij})\) changes with \(\zeta\). \(\square\)

\subsection{Conditions and proof of Theorem~\ref{thm:cluster}}
\label{app:A-7}

We state the conditions for Theorem~\ref{thm:cluster} and prove its limiting distribution and the consistency of its covariance estimator. The design of cluster \(g\) is \(\mathcal D_g=(\mathcal I_g,\{(a_t,b_t,k_t,L_t,F_t):t\in\mathcal I_g\})\), which specifies its records, displayed candidate pairs, judges and covariates. Let \(\mathcal L_g(\theta)=\sum_{t\in\mathcal I_g}\{\log(1+e^{\eta_t(\theta)})-Y_t\eta_t(\theta)\}\), so \(\Psi_g(\theta)=-\nabla_\theta\mathcal L_g(\theta)\). For the asymptotic argument, write the penalty coefficient in Equation~\eqref{eq:4} as \(\lambda=\lambda_G\). The full objective is

\[
\mathcal L_N(\theta)=\sum_{g=1}^{G}\mathcal L_g(\theta)+\frac{\lambda_G}{2}\lVert\theta\rVert^2.
\]

Theorem~\ref{thm:cluster} assumes the following conditions.

\begin{enumerate}
\tightlist
\item\label{cond:1}
  \textbf{Sampling.} Prompt clusters are independent and identically distributed, each with at most \(\bar n<\infty\) records. The sets of candidates and judges, as well as the parameter dimension, remain fixed as the number of prompts grows.
\item\label{cond:2}
  \textbf{Identification.} The parameter space \(\Theta\) is compact, \(\theta^*\) lies in its interior, and \(\theta^*\) is the unique minimizer of \(E[\mathcal L_g(\theta)]\) over \(\Theta\).
\item\label{cond:3}
  \textbf{Mean model.} For every \(t\in\mathcal I_g\), \(E[Y_t\mid\mathcal D_g]=\sigma\{\eta_t(\theta^*)\}\), with \(\eta_t\) given by Equation~\eqref{eq:2}.
\item\label{cond:4}
  \textbf{Smoothness and moments.} \(\eta_t(\theta)\) is twice continuously differentiable on a neighborhood of \(\Theta\), and \(\eta_t\), \(\nabla_\theta\eta_t\) and \(\nabla_\theta^2\eta_t\) are bounded on \(\Theta\) by a function of \(\mathcal D_g\) with finite second moment, which holds for bounded covariates.
\item\label{cond:5}
  \textbf{Curvature.} \(H_{\mathrm{cl}}=E[-\nabla_\theta\Psi_g(\theta^*)]\) is positive definite and \(V_{\mathrm{cl}}=E[\Psi_g(\theta^*)\Psi_g(\theta^*)^\top]\) is finite.
\item\label{cond:6}
  \textbf{Estimator and numerical accuracy.} The selected solution satisfies \(\hat\theta\xrightarrow{p}\theta^*\) and \(\lVert\nabla_\theta\mathcal L_N(\hat\theta)\rVert=o_p(\sqrt G)\). The deterministic penalty satisfies \(\lambda_G=o(\sqrt G)\). Since consistency implies \(\hat\theta=O_p(1)\),
\end{enumerate}

\[
\sum_g\Psi_g(\hat\theta)
=\lambda_G\hat\theta-\nabla_\theta\mathcal L_N(\hat\theta)
=o_p(\sqrt G).
\]

\emph{Proof.} Let \(A_g\geq0\) be the common upper bound in condition~\ref{cond:4}, with \(E[A_g^2]<\infty\). For \(y\in[0,1]\) and \(\eta\in\mathbb R\),

\[
0\leq\log(1+e^\eta)-y\eta\leq\log 2+|\eta|.
\]

Consequently,

\[
\sup_{\theta\in\Theta}\mathcal L_g(\theta)
\leq\bar n(\log 2+A_g),
\]

which is integrable. Compactness of \(\Theta\), continuity of the loss and condition~\ref{cond:1} therefore give a uniform law of large numbers by \cite[Lemma 2.4]{newey1994}. Moreover,

\[
\sup_{\theta\in\Theta}\frac{\lambda_G}{2G}\lVert\theta\rVert^2
\longrightarrow0.
\]

The unique population minimizer in condition~\ref{cond:2} and their Theorem 2.1 therefore imply consistency for exact minimizers of Equation~\eqref{eq:4} over \(\Theta\). For the selected numerical solution, consistency is assumed separately in condition~\ref{cond:6}.

By iterated expectations and condition~\ref{cond:3},

\[
\begin{aligned}
E[\Psi_g(\theta^*)]
&=E\left[\sum_{t\in\mathcal I_g}
r_t(\theta^*)\left\{E[Y_t\mid\mathcal D_g]-\sigma(\eta_t(\theta^*))\right\}\right]\\
&=0.
\end{aligned}
\]

By conditions~\ref{cond:1} and~\ref{cond:5} and the multivariate central limit theorem,

\[
\frac1{\sqrt G}\sum_g\Psi_g(\theta^*)
\xrightarrow{d}N(0,V_{\mathrm{cl}}).
\]

Let \(\Delta=\hat\theta-\theta^*\). With probability tending to one, the segment from \(\theta^*\) to \(\hat\theta\) lies in the interior of \(\Theta\) because \(\theta^*\) is interior and \(\hat\theta\) is consistent. Define

\[
\begin{aligned}
\mathcal H_G(\theta)&=-\frac1G\sum_g\nabla_\theta\Psi_g(\theta),\\
\mathcal H(\theta)&=E[-\nabla_\theta\Psi_g(\theta)],\\
\widetilde H_G&=\int_0^1\mathcal H_G(\theta^*+\tau\Delta)\,d\tau.
\end{aligned}
\]

The fundamental theorem of calculus applied to \(\sum_g\Psi_g(\theta)\) gives

\[
\sum_g\Psi_g(\hat\theta)
=\sum_g\Psi_g(\theta^*)-G\widetilde H_G\Delta.
\]

Using the same upper bound \(A_g\), Euclidean vector norms and the corresponding matrix operator norm, we obtain the following bounds from \(|Y_t-p_t(\theta)|\leq1\), \(p_t(\theta)\{1-p_t(\theta)\}\leq1/4\) and \(|\mathcal I_g|\leq\bar n\):

\[
\begin{aligned}
\sup_{\theta\in\Theta}\lVert\Psi_g(\theta)\rVert
&\leq\bar n A_g,\\
\sup_{\theta\in\Theta}\lVert\nabla_\theta\Psi_g(\theta)\rVert
&\leq\bar n(A_g^2/4+A_g),\\
\sup_{\theta\in\Theta}\lVert\Psi_g(\theta)\Psi_g(\theta)^\top\rVert
&\leq\bar n^2A_g^2.
\end{aligned}
\]

These upper bounds are integrable and justify the uniform laws of large numbers used below.

By Lemma 2.4 of \cite{newey1994}, we have \(\sup_{\theta\in\Theta}\lVert\mathcal H_G(\theta)-\mathcal H(\theta)\rVert\xrightarrow{p}0\), and \(\mathcal H\) is continuous. Since \(\mathcal H(\theta^*)=H_{\mathrm{cl}}\),

\[
\lVert\widetilde H_G-H_{\mathrm{cl}}\rVert
\leq\sup_{\theta\in\Theta}\lVert\mathcal H_G(\theta)-\mathcal H(\theta)\rVert
+\sup_{0\leq\tau\leq1}
\lVert\mathcal H(\theta^*+\tau\Delta)-\mathcal H(\theta^*)\rVert
=o_p(1).
\]

The first term vanishes by the uniform law of large numbers, and the second by consistency and continuity. Because \(H_{\mathrm{cl}}\) is positive definite, \(\widetilde H_G\) is invertible with probability tending to one. Rearranging the score identity and using condition~\ref{cond:6},

\[
\sqrt G\,\Delta
=\widetilde H_G^{-1}\left\{
\frac1{\sqrt G}\sum_g\Psi_g(\theta^*)
-\frac1{\sqrt G}\sum_g\Psi_g(\hat\theta)\right\}
=\widetilde H_G^{-1}\frac1{\sqrt G}\sum_g\Psi_g(\theta^*)+o_p(1).
\]

Slutsky\textquotesingle s theorem gives the limit in Theorem~\ref{thm:cluster}.

The same uniform law of large numbers applies to \(\Psi_g(\theta)\Psi_g(\theta)^\top\) under the second-moment bound in condition~\ref{cond:4}. Define

\[
\mathcal V_G(\theta)=\frac1G\sum_g\Psi_g(\theta)\Psi_g(\theta)^\top,
\qquad
\mathcal V(\theta)=E[\Psi_g(\theta)\Psi_g(\theta)^\top].
\]

The integrable upper bound established above also makes \(\mathcal V\) continuous. Since \(\mathcal V(\theta^*)=V_{\mathrm{cl}}\) and \(\hat\theta\in\Theta\) with probability tending to one,

\[
\lVert \mathcal V_G(\hat\theta)-V_{\mathrm{cl}}\rVert
\leq\sup_{\theta\in\Theta}\lVert \mathcal V_G(\theta)-\mathcal V(\theta)\rVert
+\lVert \mathcal V(\hat\theta)-\mathcal V(\theta^*)\rVert
\xrightarrow{p}0.
\]

The first term vanishes by uniform convergence and the second by consistency and continuity. Condition 6 also gives

\[
\overline{\widehat\Psi}
=G^{-1}\sum_g\widehat\Psi_g
=o_p(G^{-1/2}),
\]

so the centering term in \(\hat V/G\) vanishes. Combining these facts with the uniform convergence of \(\mathcal H_G\) and \(\lambda_G/G\to0\) gives

\[
\begin{aligned}
\frac{\hat H}{G}
&=-\frac1G\sum_g\nabla_\theta\Psi_g(\hat\theta)+\frac{\lambda_G}{G}I
\xrightarrow{p}H_{\mathrm{cl}},\\
\frac{\hat V}{G}
&=\frac{G}{G-1}\left\{
\frac1G\sum_g\widehat\Psi_g\widehat\Psi_g^\top
-\overline{\widehat\Psi}\overline{\widehat\Psi}^\top\right\}
\xrightarrow{p}V_{\mathrm{cl}}.
\end{aligned}
\]

Consequently,

\[
G\hat\Sigma
=(\hat H/G)^{-1}(\hat V/G)(\hat H/G)^{-\top}
\xrightarrow{p}H_{\mathrm{cl}}^{-1}V_{\mathrm{cl}}H_{\mathrm{cl}}^{-\top}.
\]

For the log-odds target \(h_{ij}\) in Equation~\eqref{eq:3}, let

\[
\Omega=H_{\mathrm{cl}}^{-1}V_{\mathrm{cl}}H_{\mathrm{cl}}^{-\top},
\qquad
\Omega_h=\{\nabla_\theta h_{ij}(\theta^*)\}^\top
\Omega\nabla_\theta h_{ij}(\theta^*).
\]

Suppose \(\Omega_h>0\). The delta method and covariance consistency give

\[
\sqrt G(\hat h_{ij}-h_{ij}^*)\xrightarrow{d}N(0,\Omega_h),
\qquad
G\,\widehat{\mathrm{se}}_{ij}^{\,2}\xrightarrow{p}\Omega_h,
\]

where \(\widehat{\mathrm{se}}_{ij}\) is defined in Equation~\eqref{eq:10}. Slutsky\textquotesingle s theorem then gives

\[
\frac{\hat h_{ij}-h_{ij}^*}{\widehat{\mathrm{se}}_{ij}}
\xrightarrow{d}N(0,1).
\]

Because \(\sigma\) is strictly increasing, \(q_{ij}^*\) lies between the transformed endpoints exactly when \(h_{ij}^*\) lies between the original log-odds endpoints. Thus the interval in Equation~\eqref{eq:10} has asymptotically valid coverage. \(\square\)

\subsection{Gradient of the target}
\label{app:A-8}

We derive the gradient of \(h_{ij}\) used to construct the intervals in Equation~\eqref{eq:10}. Since \(h_{ij}=\bar\gamma(s_i-s_j)\), \(s=Cu\) and \(\gamma_k=e^{v_k}\) for \(k\ne\mathrm{ref}\), we have \(\partial\bar\gamma/\partial v_k=\gamma_k/K\). Thus

\[
\nabla_\theta h_{ij}(\theta)
=\begin{pmatrix}
\bar\gamma(C_i-C_j)^\top\\[2mm]
\left((s_i-s_j)\gamma_k/K\right)_{k\ne\mathrm{ref}}\\[2mm]
0_{\dim(b)}
\end{pmatrix}.
\]

\section{Estimation and implementation}
\label{app:B}

\subsection{Parametrization, constraints and penalty}
\label{app:B-1}

We use \(s=Cu\), where \(C\) contains normalized Helmert contrasts:

\[
C_{ij}=
\begin{cases}
1/\sqrt{j(j+1)}, & 1\leq i\leq j,\\
-j/\sqrt{j(j+1)}, & i=j+1,\\
0, & i>j+1,
\end{cases}
\qquad i=1,\ldots,M,\quad j=1,\ldots,M-1.
\]

Each column sums to zero and has unit norm, and distinct columns are orthogonal, so \(C^\top C=I\) and \(C^\top\mathbf 1=0\). Non-reference sensitivities are \(\gamma_k=\exp(v_k)\) with \(v_k\in[-15,15]\). The real-data reference judge is Qwen3-235B-A22B-Instruct-2507, with \(\gamma_{\mathrm{ref}}=1\). Covariate and tie coefficients are unconstrained. The penalty is \(\tfrac{0.05}{2}\lVert\theta\rVert^2\) on every free coordinate of the summed loss, without rescaling by sample size.

\subsection{Judge-specific covariate columns and model families}
\label{app:B-2}

Within each judge, position, length and family columns are considered in that order. A column is retained only if it increases numerical rank at tolerance \(10^{-10}\); otherwise its coefficient is fixed at zero. Selection is repeated on each training or bootstrap sample.

Cross-validation and bootstrap fits use the dataset-wide length scale \(s_L\) defined in Section~\ref{sec:3-2}. Family rules are applied to lower-cased model names with organization prefixes retained.

\subsection{Optimization, convergence and identification checks}
\label{app:B-3}

Fits use L-BFGS-B with analytic gradients. Real-data fits use two starts, zero and a draw from \(N(0,0.08^2I)\). We retain the fit with the smaller objective. Simulations use the zero start. For the joint logistic-mean fits, we check:

\begin{enumerate}
\tightlist
\item
  \textbf{Stationarity:} \(\lVert\nabla\mathcal L_N(\hat\theta)\rVert_\infty/N\le10^{-7}\).
\item
  \textbf{Identification:} \(\widehat R^\top\widehat W\widehat R\), where \(\widehat R\) has rows \(\hat r_t^\top=\{\nabla_\theta\eta_t(\hat\theta)\}^\top\) and \(\widehat W=\mathrm{diag}\{\hat p_t(1-\hat p_t)\}\), has \(\dim(\theta)\) eigenvalues above the larger of \(10^{-12}\) and \(10^{-10}\) times its largest eigenvalue.
\item
  \textbf{Curvature:} the exact penalized Hessian has smallest eigenvalue above the larger of \(10^{-12}\) and \(10^{-13}\) times its largest eigenvalue, and no \(v_k\) is at either bound, −15 or 15. Covariances are computed only when this check passes.
\item
  \textbf{Cluster count:} \(G>\dim(\theta)\) is required for the full-vector cluster inference.
\end{enumerate}

All full-data fits of J, J+P and OSCAR passed these checks.

The model-based intervals in Section~\ref{sec:7} use the inverse penalized expected information:

\[
\hat\Sigma_{\mathrm{MB}}
=
\left[
\sum_t
\hat p_t(1-\hat p_t)\hat r_t\hat r_t^\top
+\lambda I
\right]^{-1}.
\]

We apply the target gradient and log-odds transformation in Equation~\eqref{eq:10} using this covariance.

\section{Data, preprocessing and protocol}
\label{app:C}

\subsection{Sources and outcomes}
\label{app:C-1}

We use the four LLM-judge datasets released with \cite{xu2026a}. UltraFeedback builds on \cite{cui2024}. MT-Bench and Chatbot Arena use the MT-Bench questions and the Chatbot Arena conversations released by \cite{zheng2023}; \cite{chiang2024} describe the Chatbot Arena platform. Verdicts for slot A, slot B and ties are coded as 1, 0 and 1/2; missing or invalid verdicts are removed, including 161 of 36,000 In-House records.

\begin{table}[!htbp]
\centering
\small
\setlength{\tabcolsep}{4pt}
\renewcommand{\arraystretch}{1.17}
\begin{tabular}{>{\raggedright\arraybackslash}p{\dimexpr 0.240\linewidth-2\tabcolsep\relax}>{\raggedright\arraybackslash}p{\dimexpr 0.200\linewidth-2\tabcolsep\relax}>{\raggedright\arraybackslash}p{\dimexpr 0.200\linewidth-2\tabcolsep\relax}>{\raggedright\arraybackslash}p{\dimexpr 0.180\linewidth-2\tabcolsep\relax}>{\raggedright\arraybackslash}p{\dimexpr 0.180\linewidth-2\tabcolsep\relax}}
\toprule
\textbf{Dataset} & \textbf{Judgments} & \textbf{Prompt clusters} & \textbf{Candidates} & \textbf{Judges} \\
\midrule
In-House & 35,839 & 990 & 45 & 18 \\
UltraFeedback & 9,726 & 9,010 & 17 & 20 \\
MT-Bench & 9,706 & 80 & 6 & 20 \\
Chatbot Arena & 9,937 & 8,691 & 20 & 10 \\
\bottomrule
\end{tabular}
\caption{Analysis sample sizes. Total 65,208 judgments.}
\label{tab:C1}
\end{table}

\subsection{Clusters and preprocessing}
\label{app:C-2}

In-House contains 45 candidate models and 990 distinct questions. The \(\binom{45}{2}=990\) unordered candidate pairs are in one-to-one correspondence with the questions: each question compares responses from one fixed pair of models, and each pair appears in only one question. We keep all judges, repetitions and display orders of a question in the same fold, so question-based splits also hold out candidate pairs. Elsewhere, clusters are released prompt identifiers.

\subsection{Matched swaps}
\label{app:C-3}

We match each In-House judgment to the released question and response texts in record order and verify agreement in judge, verdict and candidate pair. An exact-swap group contains the same judge, question and two response texts observed in both orders. There are 7,114 groups covering 22,816 records and 987 questions.

\subsection{Position summaries}
\label{app:C-4}

For judge \(k\), let \(\bar Y_k\) be the mean slot-A score across that judge\textquotesingle s valid judgments. The all-response slot-A minus slot-B rate is \(\hat D_k^{\mathrm{all}}=2\bar Y_k-1\). For a matched group \(m\), with mean slot-A scores \(\bar Y_m^{AB}\) and \(\bar Y_m^{BA}\) in the two orders, define \(d_m=\bar Y_m^{AB}+\bar Y_m^{BA}-1\) and average equally across groups. The 95\% intervals use question-cluster standard errors and \(t_{G_k-1}\) critical values, where \(G_k\) is the number of questions represented in the corresponding sample for judge \(k\).

Table~\ref{tab:C2} pairs the published sensitivity estimates from \cite[Table 2, Section 5.4]{xu2026a}, fitted on their selected eight-judge pool, with each judge\textquotesingle s all-response absolute A/B imbalance. For each judge, we use all valid In-House judgments in the analysis sample, preserving the displayed slots and weighting every judgment equally. These eight judges contribute 15,739 of the 35,839 judgments in Table~\ref{tab:C1}.

The Spearman correlation between the eight published sensitivities and \(\lvert\hat D_k^{\mathrm{all}}\rvert\) is −0.98.

\begin{table}[!htbp]
\centering
\small
\setlength{\tabcolsep}{4pt}
\renewcommand{\arraystretch}{1.17}
\begin{tabular}{>{\raggedright\arraybackslash}p{\dimexpr 0.490\linewidth-2\tabcolsep\relax}>{\raggedright\arraybackslash}p{\dimexpr 0.190\linewidth-2\tabcolsep\relax}>{\raggedright\arraybackslash}p{\dimexpr 0.140\linewidth-2\tabcolsep\relax}>{\raggedright\arraybackslash}p{\dimexpr 0.180\linewidth-2\tabcolsep\relax}}
\toprule
\textbf{Judge} & \textbf{Published sensitivity} & \textbf{Judgments} & \textbf{\(\lvert\hat D_k^{\mathrm{all}}\rvert\)} \\
\midrule
kimi-k2-thinking-turbo & 3.511 & 2,027 & 0.0784 \\
moonshot-v1-128k & 2.914 & 1,884 & 0.1178 \\
Qwen3-235B-A22B-Instruct-2507 & 2.844 & 2,007 & 0.1256 \\
Llama-4-Scout-17B-16E & 1.872 & 2,014 & 0.6311 \\
Qwen2.5-7B-Instruct-Turbo & 1.419 & 2,057 & 0.3277 \\
gemma-3n-E4B-it & 0.627 & 1,905 & 0.6378 \\
Mixtral-8x7B-Instruct-v0.1 & 0.447 & 1,947 & 0.9497 \\
DeepSeek-R1-Distill-Llama-70B & 0.046 & 1,898 & 0.9831 \\
\bottomrule
\end{tabular}
\caption{Published sensitivities and all-response absolute slot imbalance for the eight In-House judges in \cite[Table 2]{xu2026a}. Sensitivities are on the scale reported by \cite{xu2026a}.}
\label{tab:C2}
\end{table}

\section{Sensitivity experiments}
\label{app:D}

\subsection{Sample splits}
\label{app:D-1}

The calibration, refit and test splits contain 594, 198 and 198 questions and 21,214, 7,358 and 7,267 valid judgments, respectively. All records from a question remain in the same split.

\subsection{Conditional position estimation and Jeffreys correction}
\label{app:D-2}

Within an exact-swap group \(m\) containing \(n_m\) judgments from judge \(k\), choose one response, let \(Z_{mt}\) indicate its selection and let \(o_{mt}=+1\) or \(-1\) according to whether it is shown first or second. In this subsection, \(t=1,\ldots,n_m\) indexes judgments within group \(m\). The working model is

\begin{equation}
\label{eq:D1}
\mathrm{logit}\,P(Z_{mt}=1)=\alpha_m+\beta_k o_{mt}.
\end{equation}

Here \(\alpha_m\) is a group-specific response-quality intercept capturing the judge\textquotesingle s baseline preference for the chosen response, and \(\beta_k=\beta_{\mathrm{pos},k}\).

For position calibration, we assume the conditional distribution in Equation \eqref{eq:D2}, which follows from conditionally independent Bernoulli verdicts within each exact-swap group given its intercept and displayed orders. Write \(\mathbf Z_m=(Z_{mt})_{t=1}^{n_m}\) and \(\mathbf o_m=(o_{mt})_{t=1}^{n_m}\). Conditional independence under this working model gives, at \(\beta_k=\beta\) and for \(\mathbf z\in\{0,1\}^{n_m}\),

\[
P_\beta(\mathbf Z_m=\mathbf z\mid\alpha_m,\mathbf o_m)
=\frac{\exp\!\left\{\alpha_m\sum_tz_t+\beta\sum_to_{mt}z_t\right\}}
{\prod_t\{1+\exp(\alpha_m+\beta o_{mt})\}}.
\]

Let \(T_m=\sum_tZ_{mt}\) and

\[
\mathcal Y_m(T_m)
=\left\{\mathbf y\in\{0,1\}^{n_m}:\sum_ty_t=T_m\right\}.
\]

Conditional on \(T_m\), the factor involving \(\alpha_mT_m\) and the product denominator are common to all admissible configurations and cancel. For \(\mathbf z\in\mathcal Y_m(T_m)\), this gives

\begin{equation}
\label{eq:D2}
P_\beta(\mathbf Z_m=\mathbf z\mid T_m,\mathbf o_m)
=\frac{\exp\{\beta S_m(\mathbf z)\}}
{\displaystyle\sum_{\mathbf y\in\mathcal Y_m(T_m)}\exp\{\beta S_m(\mathbf y)\}},
\qquad S_m(\mathbf z)=\sum_to_{mt}z_t.
\end{equation}

Groups with \(T_m=0\) or \(T_m=n_m\) provide no conditional information: the same response is always selected. Of 4,201 calibration groups, 2,635 are informative. Always selecting the same slot after a swap instead selects different responses and can make the unpenalized position estimate diverge to \(+\infty\) or \(-\infty\), i.e., conditional separation.

For the summed conditional log-likelihood \(\ell_{c,k}\) and information \(I_{c,k}(\beta)=-\ell_{c,k}''(\beta)\), we use the Jeffreys-prior penalty of \cite{firth1993},

\begin{equation}
\label{eq:D3}
\hat\beta_k=\arg\max_\beta
\left\{\ell_{c,k}(\beta)+\frac12\log I_{c,k}(\beta)\right\}.
\end{equation}

\textbf{Existence of a finite estimate.} Define

\[
\begin{aligned}
A_m(\beta)
&=\log\sum_{\mathbf y\in\mathcal Y_m(T_m)}e^{\beta S_m(\mathbf y)},\\
\ell_m(\beta)
&=\beta S_m(\mathbf Z_m)-A_m(\beta).
\end{aligned}
\]

Differentiating the finite sum gives

\[
-\ell_m''(\beta)=A_m''(\beta)
=\mathrm{Var}_\beta\{S_m(\mathbf Z_m)\mid T_m,\mathbf o_m\}.
\]

The information \(I_{c,k}(\beta)\) is the sum of these variances over judge \(k\)\textquotesingle s exact-swap groups. Suppose at least one such group contains both display orders and has \(0<T_m<n_m\). In that group, exchanging a selected and an unselected record with opposite orders preserves \(T_m\) and changes \(S_m\) by \(2\) in absolute value. Thus two admissible configurations have different values of \(S_m\). Both have positive probability at every finite \(\beta\), so this group\textquotesingle s variance is positive and \(I_{c,k}(\beta)>0\).

Each conditional support is finite. As \(\beta\to+\infty\), probability concentrates on configurations with the largest \(S_m\), and as \(\beta\to-\infty\), it concentrates on those with the smallest \(S_m\). Each variance therefore tends to zero. Summing over the finitely many groups gives

\[
I_{c,k}(\beta)\longrightarrow0
\qquad\text{as }|\beta|\to\infty.
\]

Since each conditional probability is at most one, \(\ell_{c,k}(\beta)\leq0\). Consequently,

\[
\ell_{c,k}(\beta)+\tfrac12\log I_{c,k}(\beta)
\longrightarrow-\infty
\qquad\text{as }|\beta|\to\infty.
\]

The objective is continuous on \(\mathbb R\) and therefore attains its maximum at a finite parameter value. This establishes existence, including under conditional separation.

\subsection{Sensitivity refits, bootstrap and irregular estimates}
\label{app:D-3}

The two refits in Equation~\eqref{eq:11} minimize summed Bernoulli negative log-likelihood plus \(0.05\gamma^2/2\), subject to \(\gamma\geq0\). Both hold the calibration scores fixed, and the position-adjusted refit also holds the calibrated position effect fixed. A paired prompt bootstrap with 500 replications resamples questions within each fixed split and re-estimates calibration scores, position effects and refit sensitivities. Intervals use the 2.5\% and 97.5\% quantiles of valid draws.

GLM-4.5-Air and DeepSeek-R1-Distill-Llama-70B have conditional separation in the position calibration, and their refits give \(\widehat\gamma^{\mathrm N}_k=0\) and \(\widehat\gamma^{\mathrm P}_k=0\), respectively; Mixtral-8x7B has a boundary estimate \(\widehat\gamma^{\mathrm P}_k=0\). These three judges are reported in Table~\ref{tab:D2} and omitted from Figure~\ref{fig:1}(c).

\subsection{Joint fits and swap-and-average}
\label{app:D-4}

Full-data fits use centered scores and \(\kappa_k=\gamma_k\,\mathrm{RMS}(s)\), which is invariant to reciprocal rescaling of scores and sensitivities. Table~\ref{tab:D1} compares \(\widehat\kappa_k^{\mathrm J}/\widehat\kappa_k^{\mathrm{J+P}}\) with \(1-(\hat D_k^{\mathrm{J+P}})^2\) for both the 15 judges with finite, positive ratios and all 18 judges on In-House, and for all judges in the other datasets, where \(\hat D_k^{\mathrm{J+P}}=\tanh(\hat\beta^{\mathrm{J+P}}_{\mathrm{pos},k}/2)\) is computed from the full-data J+P fit.

For the judges included in each row, the median absolute log discrepancy is

\[
\mathrm{median}_k
\left|
\log\frac{\widehat\kappa_k^{\mathrm J}}{\widehat\kappa_k^{\mathrm{J+P}}}
-\log\left\{1-(\widehat D_k^{\mathrm{J+P}})^2\right\}
\right|.
\]

\begin{table}[!htbp]
\centering
\small
\setlength{\tabcolsep}{4pt}
\renewcommand{\arraystretch}{1.17}
\begin{tabular}{>{\raggedright\arraybackslash}p{\dimexpr 0.300\linewidth-2\tabcolsep\relax}>{\raggedright\arraybackslash}p{\dimexpr 0.170\linewidth-2\tabcolsep\relax}>{\raggedright\arraybackslash}p{\dimexpr 0.530\linewidth-2\tabcolsep\relax}}
\toprule
\textbf{Dataset} & \textbf{Judges} & \textbf{Median absolute log discrepancy} \\
\midrule
In-House, 15 judges with finite, positive ratios & 15 & 0.0044 \\
In-House, all judges & 18 & 0.0108 \\
UltraFeedback & 20 & 0.0445 \\
MT-Bench & 20 & 0.0580 \\
Chatbot Arena & 10 & 0.0107 \\
\bottomrule
\end{tabular}
\caption{Agreement between full-data sensitivity ratios and position-based predictions.}
\label{tab:D1}
\end{table}

The swap-and-average baseline fits J to preferences averaged across the two orders within each exact-swap group defined in Appendix~\ref{app:C-3}. These groups contain 22,816 In-House records, or 64\% of the analysis sample.

\subsection{Per-judge estimates}
\label{app:D-5}

Table~\ref{tab:D2} reports all 18 judges, ordered by \(\hat D_k\); intervals use the paired prompt bootstrap in Appendix~\ref{app:D-3}. Sensitivity ratios and their intervals are shown in Figure~\ref{fig:1}(c).

\begin{table}[!htbp]
\centering
\small
\setlength{\tabcolsep}{4pt}
\renewcommand{\arraystretch}{1.17}
\begin{tabular}{>{\raggedright\arraybackslash}p{\dimexpr 0.370\linewidth-2\tabcolsep\relax}>{\raggedright\arraybackslash}p{\dimexpr 0.100\linewidth-2\tabcolsep\relax}>{\raggedright\arraybackslash}p{\dimexpr 0.120\linewidth-2\tabcolsep\relax}>{\raggedright\arraybackslash}p{\dimexpr 0.120\linewidth-2\tabcolsep\relax}>{\raggedright\arraybackslash}p{\dimexpr 0.290\linewidth-2\tabcolsep\relax}}
\toprule
\textbf{Judge} & \textbf{\(\hat D_k\)} & \textbf{\(\widehat\gamma^{\mathrm N}_k\)} & \textbf{\(\widehat\gamma^{\mathrm P}_k\)} & \textbf{\(\widehat\gamma^{\mathrm P}_k-\widehat\gamma^{\mathrm N}_k\) (95\% CI)} \\
\midrule
GLM-4.5-Air-FP8* & 0.940 & 0.000 & 0.105 & 0.105 \\
DeepSeek-R1-Distill-Llama-70B* & 0.938 & 0.101 & 0.000 & −0.101 \\
Mixtral-8x7B-Instruct-v0.1* & 0.902 & 0.034 & 0.000 & −0.034 \\
gemma-3n-E4B-it & 0.788 & 0.289 & 0.610 & 0.321 (0.107, 0.575) \\
Llama-3.3-70B-Instruct-Turbo & 0.475 & 0.353 & 0.415 & 0.062 (−0.033, 0.152) \\
Qwen3-Next-80B-A3B-Instruct & 0.357 & 0.625 & 0.713 & 0.089 (0.030, 0.170) \\
Qwen2.5-7B-Instruct-Turbo & 0.357 & 0.202 & 0.235 & 0.033 (−0.019, 0.088) \\
Qwen3-235B-A22B-Instruct-2507 & 0.257 & 1.035 & 1.067 & 0.032 (−0.008, 0.094) \\
moonshot-v1-128k & 0.252 & 1.095 & 1.143 & 0.048 (−0.007, 0.130) \\
gpt-oss-20b & 0.211 & 0.760 & 0.789 & 0.029 (−0.007, 0.081) \\
gpt-oss-120b & 0.178 & 0.832 & 0.862 & 0.031 (0.002, 0.076) \\
Llama-4-Maverick-17B-128E & 0.154 & 0.654 & 0.681 & 0.028 (0.002, 0.070) \\
moonshot-v1-32k & 0.142 & 0.962 & 0.982 & 0.021 (0.000, 0.061) \\
kimi-k2-0905-preview & 0.124 & 0.920 & 0.922 & 0.002 (−0.016, 0.035) \\
kimi-k2-thinking-turbo & 0.122 & 0.931 & 0.960 & 0.028 (0.002, 0.077) \\
deepseek-chat & −0.058 & 0.646 & 0.642 & −0.004 (−0.017, 0.009) \\
cogito-v2-preview-llama-109B & −0.405 & 0.351 & 0.409 & 0.057 (−0.016, 0.148) \\
Llama-4-Scout-17B-16E & −0.711 & 0.356 & 0.672 & 0.316 (0.181, 0.531) \\
\bottomrule
\end{tabular}
\caption{Per-judge results for Section~\ref{sec:6-2}, with pointwise 95\% percentile intervals. Asterisks mark the three judges with zero or undefined sensitivity ratios described in Appendix~\ref{app:D-3}.}
\label{tab:D2}
\end{table}

\section{Held-out model comparison}
\label{app:E}

\subsection{Models compared}
\label{app:E-1}

\begin{table}[!htbp]
\centering
\small
\setlength{\tabcolsep}{4pt}
\renewcommand{\arraystretch}{1.17}
\begin{tabular}{>{\raggedright\arraybackslash}p{\dimexpr 0.220\linewidth-2\tabcolsep\relax}>{\raggedright\arraybackslash}p{\dimexpr 0.240\linewidth-2\tabcolsep\relax}>{\raggedright\arraybackslash}p{\dimexpr 0.540\linewidth-2\tabcolsep\relax}}
\toprule
\textbf{Model} & \textbf{Judge sensitivity \(\gamma_k\)} & \textbf{Judge-specific terms} \\
\midrule
J & learned & none \\
J+P & learned & position \\
J+L & learned & length \\
J+F & learned & family \\
BT+PLF & fixed at 1 & position, length, family \\
Davidson+PLF & fixed at 1 & position, length, family; tie parameter \(\tau_k\) \\
OSCAR & learned & position, length, family \\
\bottomrule
\end{tabular}
\caption{Models compared in Table~\ref{tab:1}. Except for Davidson+PLF, all minimize Equation~\eqref{eq:4} with ties coded as 1/2 and predict \(\hat p_t=\sigma(\hat\eta_t)\).}
\label{tab:E1}
\end{table}

Davidson+PLF uses \(\eta_t=s_{a_t}-s_{b_t}+\beta_{\mathrm{pos},k_t}+\beta_{\mathrm{len},k_t}L_t+\beta_{\mathrm{fam},k_t}F_t\) and the tie model of \cite{davidson1970},

\begin{equation}
\label{eq:E1}
\bigl(P(A),P(B),P(\mathrm{tie})\bigr)
=\mathrm{softmax}\bigl(\eta_t/2,-\eta_t/2,\tau_{k_t}\bigr).
\end{equation}

It minimizes the three-outcome negative log-likelihood with the same ridge on all free coordinates, including \(\tau_k\). Its prediction for the common evaluation metric is \(\hat p_t=\hat P(A)+\tfrac12\hat P(\mathrm{tie})\).

\subsection{Held-out evaluation}
\label{app:E-2}

We randomly partition prompt clusters into three folds. Each model is trained on two folds and evaluated on the third, using the same fold assignments and covariate definitions across methods. Pooled out-of-fold predictions are clipped to \([10^{-15},1-10^{-15}]\) and scored by the mean log loss

\begin{equation}
\label{eq:E2}
\mathrm{LL}=\frac1N\sum_{t=1}^N
\{-Y_t\log\hat p_t-(1-Y_t)\log(1-\hat p_t)\}.
\end{equation}

This is a proper score for the common win probability \(E(Y_t)=P(A)+\frac{1}{2}P(\mathrm{tie})\), including the mean implied by Davidson+PLF.

\subsection{Paired refit bootstrap}
\label{app:E-3}

Each bootstrap replication resamples prompt clusters with replacement separately within each fixed fold, retaining all records and multiplicities. The models in each reported contrast are refitted on the same resampled training folds and evaluated on the same resampled test folds. All model fits in the 200 bootstrap replications satisfied the stationarity criterion, and all comparison graphs were connected. We report the 2.5th and 97.5th percentiles of the 200 paired bootstrap loss differences.

\begin{table}[!htbp]
\centering
\small
\setlength{\tabcolsep}{4pt}
\renewcommand{\arraystretch}{1.17}
\begin{tabular}{>{\raggedright\arraybackslash}p{\dimexpr 0.240\linewidth-2\tabcolsep\relax}>{\raggedright\arraybackslash}p{\dimexpr 0.250\linewidth-2\tabcolsep\relax}>{\raggedright\arraybackslash}p{\dimexpr 0.140\linewidth-2\tabcolsep\relax}>{\raggedright\arraybackslash}p{\dimexpr 0.370\linewidth-2\tabcolsep\relax}}
\toprule
\textbf{Dataset} & \textbf{Difference} & \textbf{Estimate} & \textbf{Percentile 95\% CI} \\
\midrule
In-House & J − J+P & 0.1414* & (0.1369, 0.1453) \\
In-House & J+P − OSCAR & 0.0005 & (−0.0015, 0.0011) \\
In-House & BT+PLF − OSCAR & 0.0064* & (0.0046, 0.0075) \\
UltraFeedback & J+P − OSCAR & 0.0047 & (−0.0010, 0.0057) \\
UltraFeedback & BT+PLF − OSCAR & 0.0056* & (0.0003, 0.0069) \\
MT-Bench & J+P − OSCAR & 0.0003 & (−0.0118, 0.0009) \\
MT-Bench & BT+PLF − OSCAR & 0.0136* & (0.0044, 0.0178) \\
Chatbot Arena & J+P − OSCAR & 0.0070* & (0.0031, 0.0095) \\
Chatbot Arena & BT+PLF − OSCAR & 0.0149* & (0.0101, 0.0187) \\
\bottomrule
\end{tabular}
\caption{Paired held-out log-loss differences with 95\% percentile intervals. Positive differences favor the second model. An asterisk indicates that the 95\% CI excludes zero.}
\label{tab:E2}
\end{table}

\section{Human agreement on MT-Bench}
\label{app:F}

\subsection{Human votes and response matching}
\label{app:F-1}

We use 1,689 first-turn votes from the official MT-Bench human judgments \citep{zheng2023}, covering 80 prompts and the same six candidates as the LLM-judge data. Both response texts match those in the LLM-judge data. Votes for the first model, second model and either tie category are coded as 1, 0 and 1/2. LLM-judge models are fitted only to LLM judgments.

\subsection{Constructing the human reference ranking}
\label{app:F-2}

We construct the human reference ranking by fitting a Bradley--Terry model to the human votes. Each candidate \(i\) receives a score \(s_i^H\), and the model predicts \(\sigma(s_a^H-s_b^H)\) for a comparison between candidates \(a\) and \(b\). The superscript \(H\) denotes human. We estimate these scores by minimizing Equation~\eqref{eq:4}, using \(\eta_t=s_{a_t}^H-s_{b_t}^H\), the constraint \(\sum_i s_i^H=0\), and ridge coefficient \(\lambda=0.05\). Candidates are ranked in decreasing order of their fitted scores.

\subsection{Predicting human votes from LLM-based scores}
\label{app:F-3}

We evaluate how well candidate scores estimated from LLM judgments predict human votes. For each of J, J+P and OSCAR, let \(\widehat s_i\) denote the fitted score of candidate \(i\) from LLM judgments. We convert score differences into human-vote probabilities using

\begin{equation}
\label{eq:F1}
\hat p=\sigma\{\widehat\alpha_0^H+\widehat\alpha_1^H(\widehat s_a-\widehat s_b)\}.
\end{equation}

We divide the 80 prompts into five folds. For each held-out fold, we keep the candidate scores fixed and estimate the calibration intercept \(\alpha_0^H\) and slope \(\alpha_1^H\) using human votes from the other four folds, with ridge \(10^{-6}\). We then predict human votes in the held-out fold. Predictions from all five held-out folds are pooled to compute log loss using Equation \eqref{eq:E2}.

\subsection{Confidence intervals}
\label{app:F-4}

We use a prompt bootstrap to quantify uncertainty in human-vote log loss and differences between models. In each of 200 replications, we sample prompts with replacement and apply the same sampled prompt counts to both the LLM judgments and human votes. We then refit J, J+P and OSCAR, repeat the five-fold calibration in Appendix~\ref{app:F-3}, and calculate each model\textquotesingle s log loss and the differences between models. The 2.5th and 97.5th percentiles of these bootstrap results form the 95\% confidence intervals reported in Table~\ref{tab:F1}.

\begin{table}[!htbp]
\centering
\small
\setlength{\tabcolsep}{4pt}
\renewcommand{\arraystretch}{1.17}
\begin{tabular}{>{\raggedright\arraybackslash}p{\dimexpr 0.280\linewidth-2\tabcolsep\relax}>{\raggedright\arraybackslash}p{\dimexpr 0.180\linewidth-2\tabcolsep\relax}>{\raggedright\arraybackslash}p{\dimexpr 0.180\linewidth-2\tabcolsep\relax}>{\raggedright\arraybackslash}p{\dimexpr 0.180\linewidth-2\tabcolsep\relax}>{\raggedright\arraybackslash}p{\dimexpr 0.180\linewidth-2\tabcolsep\relax}}
\toprule
\textbf{Metric} & \textbf{J} & \textbf{J+P} & \textbf{OSCAR} & \textbf{Gain of J+P over J} \\
\midrule
Held-out human-vote log loss, Equation \eqref{eq:E2} & 0.5687 (0.5334, 0.5988) & 0.5604 (0.5282, 0.5912) & 0.5606 (0.5285, 0.5920) & 0.0083 (0.0022, 0.0140) \\
\bottomrule
\end{tabular}
\caption{Human-vote log loss with 95\% percentile intervals from the prompt bootstrap. Positive gain denotes lower log loss.}
\label{tab:F1}
\end{table}

\section{Simulation details}
\label{app:G}

\subsection{Simulation design and generating parameters}
\label{app:G-1}

We simulate eight candidates and five judges. Each replication generates \(G\) prompts, with all five judges evaluating the same displayed candidate pair once per prompt. This gives \(N=5G\) judgments in total. For each prompt, we draw an ordered pair of distinct candidates uniformly from the eight candidates. Candidate \(i=0,\ldots,7\) has verbosity \(\nu_i=0.8+0.25s_i^*\) for even \(i\) and \(\nu_i=-0.8+0.25s_i^*\) for odd \(i\). If the first response is less verbose, the slots are exchanged with probability 0.7, so the more verbose response is first in 85\% of prompts. After any swap, let \(a_g\) and \(b_g\) denote the candidates in the first and second slots. Every judgment \(t\) on prompt \(g\) uses \(a_t=a_g\), \(b_t=b_g\) and \(L_t=L_g\). For judgment \(t\), write \(g=g(t)\) for its prompt. With bias multiplier \(c\),

\begin{equation}
\label{eq:G1}
\begin{aligned}
L_g&=\frac{\nu_{a_g}-\nu_{b_g}+e_g}{1.5},\qquad e_g\sim N(0,0.8^2),\\
F_t&=\mathbf 1\{a_g\bmod3=k_t\bmod3\}-\mathbf 1\{b_g\bmod3=k_t\bmod3\},\\
\eta_t^*&=\gamma_{k_t}^*(s_{a_g}^*-s_{b_g}^*)
+c(\beta_{\mathrm{pos},k_t}^*+\beta_{\mathrm{len},k_t}^*L_g+\beta_{\mathrm{fam},k_t}^*F_t),\\
Y_t&=\mathbf 1\{U_t<\sigma(\eta_t^*)\},\qquad
U_t=\Phi(\sqrt{\rho}\,\varepsilon_{g(t)}^{\mathrm{cl}}+\sqrt{1-\rho}\,\varepsilon_t^{\mathrm{ind}}).
\end{aligned}
\end{equation}

Here \(\Phi\) is the standard normal distribution function, and \(\varepsilon_g^{\mathrm{cl}}\) and \(\varepsilon_t^{\mathrm{ind}}\) are standard normal variables. All noise draws, including \(e_g\), are mutually independent and independent of the candidate-pair and display-order draws. Consequently, conditional on the displayed pair and covariates, \(U_t\) is uniform on \((0,1)\), giving the logistic mean \(\sigma(\eta_t^*)\). The shared \(\varepsilon_g^{\mathrm{cl}}\) induces within-prompt dependence. The parameter \(\rho\) satisfies \(0\leq\rho\leq1\) and controls the strength of within-prompt dependence by setting the fraction of Gaussian noise variance shared across judgments. Candidate and judge families are their indices modulo 3. The true win probabilities \(q^*_{ij}\) are computed from Equation~\eqref{eq:3} using the generating scores and sensitivities. Simulation fits use Judge 0 as the reference judge, ridge coefficient \(\lambda=0.05\), and zero initialization for all free parameters.

\begin{table}[!htbp]
\centering
\small
\setlength{\tabcolsep}{4pt}
\renewcommand{\arraystretch}{1.17}
\begin{tabular}{>{\raggedright\arraybackslash}p{\dimexpr 0.270\linewidth-2\tabcolsep\relax}>{\raggedright\arraybackslash}p{\dimexpr 0.730\linewidth-2\tabcolsep\relax}}
\toprule
\textbf{Quantity} & \textbf{Value (Judge 0, Judge 1, Judge 2, Judge 3, Judge 4 for judge parameters)} \\
\midrule
Candidate scores \(s^*\) & −0.850, −0.607, −0.364, −0.121, 0.121, 0.364, 0.607, 0.850 (equally spaced) \\
Judge sensitivities \(\gamma^*\) & \(\exp(0,\,0.2,\,-0.3,\,0.35,\,-0.15)\) = 1.000, 1.221, 0.741, 1.419, 0.861 \\
Position intercepts \(\beta_{\mathrm{pos}}^*\) & 0.90, −0.70, 0.50, 1.00, −0.40 \\
Length slopes \(\beta_{\mathrm{len}}^*\) & 0.30, 0.50, 0.40, 0.35, 0.55 \\
Family coefficients \(\beta_{\mathrm{fam}}^*\) & 0.18, −0.12, 0.22, 0.10, −0.15 \\
Judge families & 0, 1, 2, 0, 1 \\
Main setting & \(c=1\), \(\rho=0.6\), \(N\in\{1000,2500,5000,10000\}\), 100 replications per \(N\) \\
\bottomrule
\end{tabular}
\caption{Generating parameters for the Section~\ref{sec:7} simulations.}
\label{tab:G1}
\end{table}

\subsection{Control settings for bias and within-prompt dependence}
\label{app:G-2}

At \(N=2500\), three controls remove bias, dependence or both, with 100 replications each. The multiplier \(c\) scales the position, length and family effects, with \(c=0\) removing all three and \(c=1\) retaining their Table~\ref{tab:G1} values. The parameter \(\rho\) controls dependence within a prompt by setting the fraction of Gaussian noise variance shared across its judgments, with \(\rho=0\) removing the shared noise so that judgments within a prompt are independent conditional on the displayed pair and covariates. All other generating and fitting settings are unchanged.

\begin{table}[!htbp]
\centering
\small
\setlength{\tabcolsep}{4pt}
\renewcommand{\arraystretch}{1.17}
\begin{tabular}{>{\raggedright\arraybackslash}p{\dimexpr 0.220\linewidth-2\tabcolsep\relax}>{\raggedright\arraybackslash}p{\dimexpr 0.070\linewidth-2\tabcolsep\relax}>{\raggedright\arraybackslash}p{\dimexpr 0.070\linewidth-2\tabcolsep\relax}>{\raggedright\arraybackslash}p{\dimexpr 0.230\linewidth-2\tabcolsep\relax}>{\raggedright\arraybackslash}p{\dimexpr 0.220\linewidth-2\tabcolsep\relax}>{\raggedright\arraybackslash}p{\dimexpr 0.190\linewidth-2\tabcolsep\relax}}
\toprule
\textbf{Control} & \textbf{\(c\)} & \textbf{\(\rho\)} & \textbf{Coverage (\%): OSCAR, cluster} & \textbf{OSCAR, model-based} & \textbf{J, cluster} \\
\midrule
No bias & 0 & 0.6 & 96.3 & 83.6 & 95.6 \\
No dependence & 1 & 0 & 95.4 & 95.0 & 38.1 \\
Neither & 0 & 0 & 93.9 & 93.9 & 93.6 \\
\bottomrule
\end{tabular}
\caption{Coverage (\%) of 95\% confidence intervals for \(q\) in the three control settings. Coverage is averaged over the 28 candidate pairs within each replication and then over 100 replications.}
\label{tab:G2}
\end{table}

\subsection{Monte Carlo uncertainty}
\label{app:G-3}

All methods are evaluated on the same simulated data within each replication. For each metric, the Monte Carlo standard error is the standard deviation of its 100 replication-level values divided by \(\sqrt{100}\). We construct 95\% Monte Carlo intervals using the \(t_{99}\) critical value. Intervals for differences between methods use paired replication-level differences.

\end{document}